%% file: main.tex
\documentclass[conference]{IEEEtran}
\IEEEoverridecommandlockouts

\input{header}
\definecolor{tblue}{RGB}{93, 142, 150}

\def\BibTeX{{\rm B\kern-.05em{\sc i\kern-.025em b}\kern-.08em
    T\kern-.1667em\lower.7ex\hbox{E}\kern-.125emX}}
\begin{document}

\title{Multi-State Brain Network Discovery}

\makeatletter
\newcommand{\linebreakand}{%
  \end{@IEEEauthorhalign}
  \hfill\mbox{}\par
  \mbox{}\hfill\begin{@IEEEauthorhalign}
}
\makeatother

\author{
  \IEEEauthorblockN{Hang Yin}
  \IEEEauthorblockA{\textit{Worcester Polytechnic Institute}\\
    Worcester, MA, USA \\
    hyin@wpi.edu}
  \and
  \IEEEauthorblockN{Yao Su}
  \IEEEauthorblockA{\textit{Worcester Polytechnic Institute}\\
    Worcester, MA, USA \\
    ysu6@wpi.edu}
  \and
  \IEEEauthorblockN{Xinyue Liu}
  \IEEEauthorblockA{\textit{Worcester Polytechnic Institute}\\
    Worcester, MA, USA \\
    xliu4@wpi.edu}
  \linebreakand 
  \IEEEauthorblockN{Thomas Hartvigsen}
  \IEEEauthorblockA{\textit{University of Virginia}\\
    Charlottesville, VA, USA \\
    hartvigsen@virginia.edu}
  \and
  \IEEEauthorblockN{Yanhua Li}
  \IEEEauthorblockA{\textit{Worcester Polytechnic Institute}\\
    Worcester, MA, USA \\
    yli15@wpi.edu}
  \and
  \IEEEauthorblockN{Xiangnan Kong}
  \IEEEauthorblockA{\textit{Worcester Polytechnic Institute}\\
    Worcester, MA, USA \\
    xkong@wpi.edu}
}

\maketitle

\begin{abstract}

Brain network discovery aims to find nodes and edges from the spatio-temporal signals obtained by neuroimaging data, such as fMRI scans of human brains.
Existing methods tend to derive representative or average brain networks, assuming observed signals are generated by only a \textit{single} brain activity state.
However, the human brain usually involves \textit{multiple} activity states, 
which jointly determine the brain activities.
The brain regions and their connectivity usually exhibit intricate patterns that are difficult to capture with only a single-state network.
Recent studies find that brain parcellation and connectivity change according to the brain activity state.
We refer to such brain networks as \textit{multi-state}, and this mixture can help us understand human behavior.
Thus, compared to a \emph{single-state} network, a \emph{multi-state} network can prevent us from losing crucial information of cognitive brain network. 
To achieve this, we propose a new model called MNGL (\underline{M}ulti-state \underline{N}etwork \underline{G}raphical \underline{L}asso), which successfully models multi-state brain networks by combining CGL (coherent graphical lasso) with GMM (Gaussian Mixture Model).
Using both synthetic and real world ADHD-200 fMRI datasets, we demonstrate that MNGL outperforms recent state-of-the-art alternatives by discovering more explanatory and realistic results.

\end{abstract}


\maketitle





\section{Introduction}
\textbf{Motivation}. Brain network discovery~\cite{BS09,ZFB10,ZWO14} is one of the most pervasive paradigms in neuroscience and involves two main tasks: \textit{brain parcellation} and \textit{edge detection}. 
In brain networks, nodes represent brain regions, and edges represent the functional/structural connections between regions.
In general, brain networks are modeled by first finding nodes that contain coherently-functioning brain regions (\ie performing \textit{brain parcellation}), then identifying edges between these nodes according to an observed sequence of brain activity (\ie \textit{edge detection}). 
Precise discovery of these networks cultivates a more refined model of the human brain. Such models become instrumental in diagnosing brain disorders~\cite{du2018classification} and analyze brain functions~\cite{van2010exploring}. Furthermore, recent studies~\cite{lin2017dynamic,anderson2018developmental,diez2018neurogenetic} have found that the difference in brain activity states can infer to distinct brain parcellations and connectivity patterns. 
Thus, an effective brain network discovery methodology must be adaptive, adjusting to the dynamism of the brain's activity state. This temporal adaptability (\ie state-based adjustment) in brain parcellation and connectivity becomes pivotal in understanding human brain networks. 
Ultimately, characterizing this mixture functional structure of brain parcellation and functional connectivity, as shown in Figure~\ref{fig:butterfly}, leads to a better understanding of brain function and human behavior.

\input{butterfly.tex}

\input{family.tex}

\textbf{Knowledge Gap}. 
Formally, the brain network discovery problem is anchored on inferring a set of functionally homogeneous brain regions as the network nodes and the mapping of their connectivity as network edges, all based on a series of brain scans taken over time.
While there are some recent solutions  \cite{liu2017collective,yin2018coherent,bai2017unsupervised} can address the problem, they often ignore the flexibility of functional network configurations. 
They assume that the brain is always in a single activity state,
implying that signals extracted from different regions of the brain at different times are members of the \textit{same} network—a notion refuted by recent studies~\cite{lin2017dynamic}.
Additionally, some studies~\cite{su2023one} have focused on obtaining parcellation by mapping the brain onto an atlas by image registration~\cite{su2022ernet,su2022abn}. The choice of atlas can influence the derived parcellation and generate distinct activity states.
However, in the context of uncertainty in mixture state assignment, identifying the appropriate atlases across time presents a significant challenge.
Thus, this paper investigates the problem of
multi-state brain network discovery, as
shown in Figure~\ref{fig:butterfly}. The goal is to design methods in brain network discovery to capture \textit{multiple} underlying brain network states, allowing for differing brain parcellation and connectivity.


\textbf{Challenges}. To incorporate the concept of \textit{multiple states} into brain network discovery, our main challenges are:
\begin{itemize}
\item \textit{Brain Network discovery}: Edge detection in functional brain network discovery focuses on direct links between the network nodes. However, the raw fMRI data usually do not contain background knowledge of node segmentation. Thus brain network discovery is the first challenge we face. 
Brain network discovery traditionally aims to use a cohesive model for inferring brain parcellation and edge detection at the same time \cite{yin2018coherent}.
However, the brain network may have different brain parcellations in different states and it is impractical to handle networks of each state with uniform node segmentation. 
Some recent brain network discovery methods \cite{liu2017collective, yin2018coherent, bai2017unsupervised} can handle learning both nodes and edges, though each has its clear limitations.
\cite{bai2017unsupervised} aims to infer brain parcellation with spatial continuity constraint for the sake of interpretability, but it fails to distinguish the direct connections and indirect connections among the network nodes. 
\cite{liu2017collective} considers the brain network discovery problem as a coherent one, they apply two separate objective functions for two sub-tasks respectively, and update each other alternatively in the same framework.
\cite{yin2018coherent} suggests CGL (Coherent Graphical Lasso) deals with coherent brain network discovery, which combines the ideas of orthogonal non-negative matrix factorization with Graphical Lasso. 
However, this method can not solve the multi-state network in Figure~\ref{fig:butterfly} due to the lack of information about state assignment.

\item \textit{Mixture of multiple brain networks}: 
Some recent works \cite{gao2016estimation, yin2020gaussian} have applied mixture models such as JGL (Joint Graphical Lasso) and MGL (Mixture Graphical Lasso) to brain network analysis. However, they all need brain parcellation to be given first.
Thus, combining the existing Gaussian mixture model with a brain network discovery method remains unsolved.

\item \textit{Dependence between brain activity states and brain network}:
Variations in brain activity states influence network estimations.
Alterations of state assignments subsequently change the outcomes of corresponding network updates.
Given this interdependency, pipeline methodologies that combine baseline techniques like CGL~\cite{yin2018coherent} and ONMtF~\cite{bai2017unsupervised} are inappropriate for multi-state brain network discovery. Because the pipeline frameworks apply the methods for brain parcellation and edge detection separately, such approaches lead to estimation inconsistency. 
\end{itemize}

\textbf{Proposed Method}. To tackle the above challenges, we propose a new model, named MNGL
, for multi-state brain network discovery, which jointly achieves brain parcellation and edge detection. 

We leverage the idea of Probabilistic Latent Semantic Analysis (PLSA)~\cite{hofmann1999probabilistic}, which was originally proposed to adopt a mixture model for natural language. PLSA assumes a given document is a mixture of topics, and the document was generated according to a probabilistic model with latent topics.
Inspired by this, first, we view brain scans as mixtures of latent states, where each state $S$ is characterized by a Gaussian distribution with its own covariance matrix $\Sigma_{\text{S}}$.
Each $\Sigma_{\text{S}}$ corresponds to a specific brain parcellation and connectivity between nodes. 
Therefore, in the generation of each brain scan, our model chooses a state $\text{S}$ based on the mode distribution $\pi$ (similar to how PLSA chooses a topic), and then generates a brain scan $B_{i} \sim \text{Multinomial}(\mathbf{0},\Sigma_{\text{S}})$ (as PLSA generates a word based on the topic chosen). 
Our model MNGL follows the basic idea of PLSA. 
By contrast, traditional brain network discovery models \cite{bai2017unsupervised, yin2018coherent} assume all brain scans are produced by a single state, which is characterized by a unified zero-mean $\Sigma$-covariance multivariate Gaussian distribution.
Thus, they are analogous to naive bayes model~\cite{mccallum1998comparison}.
Figure~\ref{fig:family} illustrates these two pairs of comparison.
To model this multi-state network, we combine CGL with GMM in a unified objective function to deal with multi-state networks. 
Compared to other mixture models \cite{gao2016estimation, yin2020gaussian},
our model only needs original brain data (a series of brain scans) as input without any prior knowledge related to nodes or their connectivity or assignments of each brain states, and outputs multiple brain networks that include both brain parcellation and connectivity structures.

\textbf{Contributions}. The contributions of this work are:
\begin{itemize}
    \item We describe the open multi-state brain network discovery problem, which is to find the underlying network structure of hybrid cognitive brain states from a series of brain scans. 
    \item We propose the first solution to this open problem, leveraging recent successes of Gaussian Mixture Models and the Coherent Graphical Lasso. 
    \item We demonstrate that our model outperforms recent state-of-the-art alternatives by discovering more accurate and realistic results on both synthetic and real fMRI datasets. 
\end{itemize}


\input{notation.tex}
\section{Preliminary}

We begin by introducing some basic models to deal with brain parcellation, edge detection, and mixture modelling.

\textbf{Brain Parcellation}. Let $\mb{X} = (\mathbf{x}_1,...,\mathbf{x}_n) \in \mathbb{R}^{p \times n}$ be the observations of a $p$-variate Gaussian distribution where $p$ denotes the number of variables and $n$ is the number of observations. Then, we let $\mb{\Sigma}$ be the covariance matrix of the $n$ samples. The Non-negative Matrix Factorization (NMF) can then be used to factorize $\mb{\Sigma}$ into two non-negative matrices:
\begin{align}
\label{NMF}
	\mb{\Sigma} \approx \mb{F}\mb{G}^\top,
\end{align}
where $\mb{F} = (\mathbf{f}_1,...,\mathbf{f}_k) \in \mathbb{R}^{p \times k}$ and $\mb{G} = (\mathbf{g}_1,...,\mathbf{g}_k) \in \mathbb{R}^{p \times k}$, and $k$ is a pre-specified number of nodes to discover. For network discovery, our target is an absolute covariance matrix $\mb{\Sigma}$. So $\mb{\Sigma}$ is a systematic analysis matrix and $\mathbf{F}$ is equal to $\mathbf{G}$, which we henceforth refer to $\mathbf{F}$ and $\mathbf{G}$ as $\mathbf{H}$. We can then extend the NMF model to weighted orthogonal non-negative factorization, or ONMtF \cite{ding2006orthogonal}, after which the objective function becomes:
\begin{equation}
\label{ONMGL}
\begin{aligned}
\min_{\mb H \geqslant 0, \mb H \mb H^T = \mb I} ||\mb{\Sigma} - \mb{HSH}^{\top}||^2.
\end{aligned}
\end{equation}
By adding non-negativity and orthogonality constraints, the model is equivalent to $k$-means clustering and the Laplacian-based spectral clustering \cite{ding2005equivalence}.


\textbf{Edge Detection}. 
As in \cite{friedman2008sparse}, directed links among the network nodes can be discovered by minimizing the following objective:
\begin{equation}
\label{Glasso}
\begin{aligned}
\min_{\mb{\Theta} \succ 0} \left(-\log \det \mb{\Theta} + \text{tr}(\mb{S}\mb{\Theta}) + \lambda ||\mb{\Theta}||_1\right),
\end{aligned}
\end{equation} 
where $\mb{S} = \frac{1}{n}\mb{XX}^T$ is the empirical covariance matrix, $\mb{\Theta}$ is the precision matrix which is the inverse of the systematic covariance matrix $\mb{\Sigma}$,
$\ell_1$ regularization is used to force sparsity, and $\mb{\lambda}$ is the parameter to control the sparseness of $\mb \Theta$. 
The edge $e_{ij}$ between $\mathbf{x}_i$ and $\mathbf{x}_j$ exists if and only if $\theta_{ij} \neq 0$, where $\theta_{ij}$ is the $(i,j)$-element of $\mb \Theta$. We prefer $\mb \Theta$ rather than matrix $S$ in ONMtF, due to the power of sparse gaussian graphic models on large-scale datasets.

\textbf{Coherent Graphical Lasso}.
The Coherent Graphical Lasso (CGL) achieves the two sub-tasks of Brain Network Discovery (node discovery and edge detection) simultaneously \cite{yin2018coherent}. CGL is a special graphical lasso with an orthogonal non-negative matrix factorization, as shown in Equation \ref{CGL}:
\begin{equation}
\label{CGL}
\begin{aligned}
& \min_{\mb{H},\mb{\Theta^{\star}}} -\log \det \mb{\Theta^{\star}} + \text{tr}(\mb{H}^\top \mb{SH} \mb{\Theta^{\star}}) + \lambda||\mb{\Theta^{\star}}||_1,\\
& \qquad \text{s.t.} \quad \mb{\Theta^{\star}} \succ 0 , \mb H \geqslant 0, \mb H \mb H^\top = \mb I\\
\end{aligned}
\end{equation}
where $\textbf{S}$ is the empirical covariance matrix, $p$ is the number of features, $k$ is the number of nodes, $\mb{\Theta^{\star}}$ is the inverse of the $k \times k$ absolute inter-node covariance matrix, and $\textbf{H}$ is a $p \times k$ cluster indicator matrix. 
However it can not be applied to the problem of the multi-state network discovery directly as it requires prior knowledge of state assignments. 

\textbf{Mixture Model}. An attractive and powerful model for multi-state problems is the Gaussian Mixture Model (GMM), where each base distribution in the mixture is a Multivariate Gaussian (MVG) with mean $\bsym{\mu}_k$ and covariance matrix $\mb{\Sigma}_k$. The probability of data sample $\bsym{x}_i$ is then
\begin{align}
    \label{eq:gmm}
    p(\bsym{x}_i \vert \bsym{\theta}) = \sum_{k=1}^{K} \phi_k \mc{N}(\bsym{x}_i \vert \bsym{\mu}_k, \mb{\Sigma}_k),
\end{align}
where $\bsym\theta$ is the model parameters, $\phi_k$ is the prior probability of the $k$-th distribution chosen to generate a sample and $\sum_{k=1}^{K} \phi_k = 1 $. Next subsection, we introduce a novel model that extends CGL into the framework of a Gaussian Mixture Model, thereby solving the multi-state brain network discovery problem.




\section{Methodology}

\textbf{Multi-State Network Graphical Lasso}. In this work, we propose the first method for Multi-State Brain Network Discovery, which we refer to as the \textbf{M}ulti-State \textbf{N}etwork \textbf{G}raphical \textbf{L}asso, or MNGL. Firstly, following the idea of \cite{yin2018coherent}, we map the original variable space $\mb X$ into a new feature space $\mb Y$ similarly on the covariance matrix $\mb \Sigma$: 
\begin{equation}
\label{map}
\begin{aligned}
& \mb {X} \leftarrow \mb {Y} = \mb{H}^\top \mb{X},\\
& \mb {\Sigma} \leftarrow \mb{\Sigma}^\star = \mb{H}^\top \mb{\Sigma} \mb{H},\\
\end{aligned}
\end{equation}
where $\mb{Y}$ denotes the new $k$-dimensional feature space where each feature represents node, $\mb{\Sigma}^\star$ represents the inter-node covariance matrix, and $\mb{H}$ represents a cluster indicator matrix. $\mb{\Sigma}^\star$ thus measures the association between each node $\mb{\bsym{y}}_{i}$ \cite{yin2018coherent}.


For the rest of this section, we describe our proposed model in terms of $\mb{\bsym{y}}_{i}$ instead of $\mb{\bsym{x}}_{i}$. $k$ is the index of nodes, $j$ is the index of distributions, $i$ is the index of samples. $\bsym\mu^\star_j$ and $\bsym\Sigma^\star_j$ represent the parameters of mean vector and covariance matrix corresponding to the $j \text{-th}$ mixture gaussian distribution of $\mb{Y}$, respectively. Then, $\bsym\Theta^\star_j$ represents the inverse matrix of covariance matrix $\bsym\Sigma^\star_j$. 
More special notations are collected in Table \ref{tab:notation}.

According to the notation above, given the number of base distributions $m$ and the number of node $k$, we assume the observed sample of target feature space can be mapped into a new feature space (nodes), which also follows a mixture of the $k$ gaussian distributions. The sample size is given as $n$.
Thus, the joint probability of these nodes $\mb{Y} = (\bsym{y}_1^\top, \cdots, \bsym{y}_n^\top) \in \mathbb{R}^{n \times k}$ is given by
\begin{align*}
      & p(\mb{Y} \vert \{\mb{\Theta}^\star_j\},  \{\bsym{\mu}^\star_j\}, \{\phi_j\}) = \prod_{i=1}^{n}\sum_{j=1}^{m}  \phi_j \mc{N}(\bsym{y}_i \vert \bsym{\mu}^\star_j, \mb{\Sigma}^\star_j ).
\end{align*}
By assuming $\bsym\mu^\star_j = \bsym{0}$ without losing generality, the negative log likelihood (NLL) in terms of $\{\mb{\Theta}^\star_k\}$ is given by,
\begin{equation}
\label{eq:mgglsimple}
\begin{aligned}
     \text{NLL}(\bsym\theta) &= - \sum_{i=1}^{n}\text{log}\Big(\sum_{j=1}^{m} \phi_j\mc{N}(\bsym{y}_i \vert \bsym{0}, (\mb{\Theta}^{\star}_{j})^{-1})\Big),
\end{aligned}
\end{equation}
where $\bsym\theta = \{\phi_1, \cdots, \phi_m, \mb{{\Theta}}^{\star}_1, \cdots, \mb{{\Theta}}^{\star}_m\}$ is the model parameters.

\textbf{Latent States}. In order to solve the Equation \ref{eq:mgglsimple}, we follow the idea of Jensen inequality and build a latent variable in the sum term of each expression in $\log$. Since there are $m$ separate latent distributions, each data sample of the corresponding node $\bsym{y}_i$ could come from one of the $K$ distributions. We therefore construct a latent variable $\mb{Q}(z_{ij})$ which we constrain such that $\sum_{j=1}^{m} \mb{Q}(z_{ij}) = 1 $.
Then, the NLL function can be rewritten as follows:
\begin{align}
\label{latent}
    \text{NLL}(\bsym\theta) &= - \sum_{i=1}^{n}\log\sum_{j=1}^{m}\Big(\frac{\mb{Q}(z_{ij}) p\big(\bsym{y}_i \vert  \mb{\Theta}^{\star}_j, \phi_j\big)}{\mb{Q}(z_{ij})}\Big)\\
    &= - \sum_{i=1}^{n}\log\sum_{j=1}^{m}\Big(\frac{p\big(\bsym{y}_i,z_{ij} \vert \mb{\Theta}^{\star}_k, \phi_j\big)}{\mb{Q}(z_{ij})}\Big).
\end{align}
We next prove that this can be treated as the posterior probability of the $i$-th observation generated by the $j$-th distribution.

According to the Jensen inequality, the expression in the Equation \ref{latent} can be rewritten for the EM algorithm to optimize the function, which can be split into expectation and maximization steps, respectively. 

\textbf{Expectation}. First, according to the Jensen inequality, we know that when the optimal function is convex, 
\begin{align}
    f(E(x)) \leqslant E(f(x)).
\end{align}
Because NLL is convex, and $\sum_{j=1}^{m}\Big(\frac{p\big(\bsym{y}_i,z_{ij} \vert \mb{{\Theta}}^{\star}_j, \phi_j\big)}{\mb{Q}(z_{ij})}\Big)$ can be treated as the expectation of $p\big(\bsym{y}_i,z_{ij} \vert \mb{{\Theta}}^{\star}_j, \phi_j\big)$. So we apply Jensen inequality here to find a lower bound:
\begin{align}
    \text{NLL}(\bsym\theta) 
    &\leqslant - \sum_{i=1}^{n} \sum_{j=1}^{m} \mb{Q}(z_{ij}) \log (p\big(\bsym{y}_i,z_{ij} \vert \mb{{\Theta}}^{\star}_j , \phi_j\big)).
\end{align}
These terms are only equal when 
\begin{align}
    \frac{p(\bsym{y}_i,z_{ij})}{\mb{Q}(z_{ij})} = C,
\end{align}
where $C$ is a constant. So, we simply have:
\begin{align}
\label{eq:equal}
    \sum_{j=1}^{m}p(\bsym{y}_i,z_{ij}) = C \sum_{j=1}^{m} \mb{Q}(z_{ij}) = C,\\
    \mb{Q}(z_{ij}) = \frac{p(\bsym{y}_i,z_{ij})}{\sum_{j=1}^{m}p(\bsym{y}_i,z_{ij})} = r_{ij}.
\end{align}
The equation of $\text{NLL}(\bsym\theta)$
is correct only when the constraint of $\mb{Q}(z_{ij})$ is true. Thus we can conclude that the latent variable is the posterior probability of the $i$-th observation generated by the $j$-th distribution. Therefore, we can compute each $r_{ij}$ based on the initialization or update results of $\mb{{\Theta}}^{\star}_{j}$ and $\mb{\phi}_j$.

\textbf{Maximization}. Given the $\bsym{r}_{ij}^{(t)}$ from the Expectation step, we update $\hat{\phi}_j$, $\hat{\mb{H}}_j$ and $\hat{\mb{\Theta}}^{\star}_j$, respectively. First, we update $\hat{\phi}_j$ based as follows:
\begin{align}
\label{eq:update_phi}
    \hat{\phi}_j^{(t)} = \frac1n\sum_{i=1}^{n} r_{ij}^{(t)}.
\end{align}
The remaining problem is to find the optimal estimations of $\mb{H}_j$ and $\mb{{\Theta}}^{\star}_j$ that maximizes the expectation we obtain in the E step. Through a simple proof, it is equivalent to minimize the following function:
\begin{equation}
\begin{aligned}
 & \min \sum_{i=1}^{n}\sum_{j=1}^{m} -r_{ij}^{(t)} \big(\text{log} \vert\mb{{\Theta}}^{\star}_j\vert - \bsym{x}_i^\top \mb{H}_j \mb{{\Theta}}^{\star}_j \mb{H}^\top_j \bsym{x}_i\big),\\
 & \qquad \text{s.t.} \quad \mb{{\Theta}^\star} \succ 0,\mb H \geqslant 0,\mb{H}\mb{H}^\top=\mb{I}.
\end{aligned}
\end{equation}
Intuitively, the problem above is equivalent to $m$ separate conventional graphical lasso sub-problems weighted by $r_{ij}^{(t)}$ where each sub-problem has the form of
\begin{equation}
\begin{aligned}
    \label{eq:subproblem}
    & \min - \log \vert \mb{{\Theta}}^{\star}_j \vert + \text{tr}( \tilde{\mb{X}}_j^\top\mb{H}_j\mb{{\Theta}}^{\star}_j\mb{H}^\top_j\tilde{\mb{X}}_j),\\
    & \qquad \text{s.t.} \quad \mb{{\Theta}}^{\star}_j \succ 0,\mb H_j \geqslant 0,\mb{H}_j\mb{H}^\top_j=\mb{I},
\end{aligned}
\end{equation}
where  $\tilde{\mb{X}}_j = (\sqrt{r_{1j}/s_j} \bsym{x}_1^\top, \cdots, \sqrt{r_{nj}/s_j} \bsym{x}_n^\top)$, $\bsym{r}_{ij} = (r_{1j}, \cdots, r_{nj})^\top$ and  $s_j = \sum_{i=1}^{n} r_{ij}$. 
Then we bring in the $\ell_1$ regularization $\lambda ||\mb{{\Theta}}^{\star}_j||_1$ to obtain the final objective function for the Maximization step.

This problem is not convex \emph{w.r.t.}  $\{\mb{{\Theta}}^{\star}_j\}$, but we could solve it alternatively for each $\mb{{\Theta}}^{\star}_j$ by regarding other $\mb{{\Theta}}^{\star}_{j'\ne j}$ fixed.
Each sub-problem of  $\mb{{\Theta}}^{\star}_j$ is exactly in the form of Equation \ref{eq:subproblem} plus the $\ell_1$ regularization terms.
Thus the estimation of $\mb{{\Theta}}^{\star}_j$ could be solved by any existing method for solving Graphical Lasso without significant modifications.
To estimate $\mb{H}_j$ we follow the algorithm similar to NMF, using Karush–Kuhn–Tucker (KKT) complementary slackness conditions to enforce the non-negativity and orthogonality constraints, then solving the estimation of $\mb{H}_j$ by the multiplicative update rule. Thus, we have:
\begin{equation}
\label{ite_H}
\begin{aligned}
(\mb{\hat{H}}^{(t+1)}_{j})_\text{ls} = \left(\mb{\hat{H}}^{(t)}_{j} \right)_\text{ls}{
\left(\frac{\tilde{\mb{X}}_j\tilde{\mb{X}}^\top_j \mb{\hat{H}}^{(t)}_j \mb{{\hat{\Theta}}^{{\star}{-}}}_j+\mb{\hat{H}}^{(t)}_j\lambda^{-}_1}{\tilde{\mb{X}}_j\tilde{\mb{X}}^\top_j\mb{\hat{H}}^{(t)}_j \mb{{\hat{\Theta}}^{{\star}{+}}}_j+\mb{\hat{H}}^{(t)}_j\lambda^{+}_1}\right)}_\text{ls}.
\end{aligned}
\end{equation}
Here $\lambda_{1}$ is $k \times k$  Lagrangian multip matrices following the non-negativity constraint and its compact expression follows as below: 
\begin{equation}
\label{lambda1_2}
\begin{aligned}
\lambda_1 =  -\mb{\hat{H}_j}^\top\tilde{\mb{X}}_j\tilde{\mb{X}}^\top_j\mb{\hat{H}_j}\mb{{\hat{\Theta}}^{\star}}_j.
\end{aligned}
\end{equation}
To make sure each part is non-negative, We divide the $\lambda_1$ and $\mb{\hat{\Theta}}^{\star}$ into two parts, respectively:
\begin{equation}
\label{divide}
\begin{aligned}
& \lambda_1 = \lambda^{+}_1 - \lambda^{-}_1,\\
& \lambda^{+}_1 = \frac{(|\lambda_1|+\lambda_1)}{2},\\
& \lambda^{-}_1 = \frac{(|\lambda_1|-\lambda_1)}{2}.\\
\end{aligned}
\end{equation}
The same is true on the $\mb{{\hat{\Theta}}^{\star}}_j$. Thus we can make sure the sign of numerator and denominator are all positive, abiding by the non-negative constraint of $\mb{H}_j$.

In each iteration of the Maximization step, the alternating optimization repeats until all estimated $\hat{\mb{\Theta}}^{\star}_j$, $\hat{\mb{H}}_j$ and $\hat{\mb{\phi}}_j$ become stable or reaches the maximal number of iterations.
The final solutions to Equation \ref{eq:subproblem} and the updated $\{\hat{\mb{\phi}}_j\}$ are obtained using Equation \ref{eq:update_phi} are used in the upcoming iteration of Expectation step to update the responsibility weights $\{r_{ij}\}$.
This looping of Expecation and Maximization repeats until the loss function converges. The MNGL algorithm is also summarized in Algorithm \ref{mggl}.

\begin{algorithm}[t]
\caption{Algorithm for \texttt{MNGL}}
\label{mggl}
\begin{algorithmic}[1]
\Require 

i: $\mb{X}$: The observations of $D$-variate Gaussian distribution

ii: $m$: the number of Gaussian distributions

iii: $k$: the number of nodes (groups)

iv: $\lambda_1$: the Lagrangian multiplier of the $\ell_1$ regularization in graphical lasso

v: $\text{iter}_\text{max}$: the maximum number of iteration

Output: $\hat{\mb{\Theta}}^{\star}_j$, $\mb{\hat{H}_j}$ and $\mb{\Hat{\phi}_j}$
\State Initialization: initialize $\hat{\mb{\phi}(0)}_j^{(0)}$, $\hat{\mb{\Theta}}^{\star (0)}_j$ ,$\hat{\mb{H}}_j^{(0)}$ and $r_{ij}^{(0)}$
\Repeat
\State E step: Update the latent variable $r_{ij}^{(t)}$ with given $\hat{\mb{\phi}}_j^{(t-1)}$, $\hat{\mb{\Theta}}^{\star (t-1)}_j$ and $\hat{\mb{H}}_j^{(t-1)}$
\State M step: Update $\hat{\mb{\phi}}_j^{(t)}$, $\hat{\mb{\Theta}}^{\star (t)}_j$ and $\hat{\mb{H}}_j^{(t)}$ with  $r_{ij}^{(t)}$
\Until{$iter={iter}_\text{max} \text{ or convergence}$}

\end{algorithmic}
\end{algorithm}



\textbf{Initialization}. As shown in Algorithm \ref{mggl}, we need to provide starting values for each estimator. The following scheme we found empirically works well in our experiments. For each observation $i = 1,\dots,n$, we distribute the observation randomly a class $j \in \{1,\dots,m\}$. Then we assign a weight $\hat{r}_{ij} = 0.9$ for this observation $i$ and distribution $k$ and $\hat{r}_{ij} = \frac{0.1}{m-1}$ for all other distributions. In the Maximization step, we update $\hat{\mb{\Theta}}^{\star}_j$ from the initial values $\hat{\mb{\Theta}}^{\star (0)}_j$ computed by CGL based on the whole samples and $\hat{\phi}_j$ from the initial values $\hat{\phi}_k = \frac{1}{m}$. Then for $\hat{\mb{H}}_j$, according to the Equation \ref{ite_H}, we note that if $(\mb{\hat{H}}^{(t+1)}_{j})_{ls} = 0$ in one iteration, it will never jump out from this local solution. Thus, our experiments we initialize $\mb{\hat{H}}_j^{(0)}$ by performing $k$-means clustering then setting $\mb{\hat{H}}_j^{(0)} \leftarrow \mb{\hat{H}}_j^{(0)}+0.2$.

\section{Empirical Study}

We begin by evaluating our method using synthetic data where we have access to the ground truth brain states. To comprehensively evaluate the proposed model, we conduct experiments to answer the following research questions: 
\begin{itemize}
\item{\textit{RQ 1}:} How does sample size affect MNGL's performance relative to state-of-the-art alternatives? 
\item{\textit{RQ 2}:} How robust is MNGL to the presence of noise compared to other recent models?
\item{\textit{RQ 3}:} How do hyper-parameters in comparative experiments impact each model's performance?
\item{\textit{RQ 4}:} How does the number of nodes affect each compared model?
\end{itemize}
\subsection{Experiment Setup}

\subsubsection{Synthetic Data with Ground-Truth}
We evaluate the performance of our model on synthetic data, where the ground-truth is known. 
The first step of generating these synthetic data is to build a mixture Gaussian distribution of network structure. By following the approach of \cite{yin2018coherent} in generating a single network, we generate $m$ different block-diagonal matrix $\mb\Theta_j$ and $\mb{H}_j$ firstly. We refer to each diagonal block as the node in real-case. 
For each $\mb\Theta_j$, we give random sparity structures for each block $\mb\Theta_{G_i,G_j}$. 
In this paper, we design each diagonal block $\mb\Theta_{G_i,G_i}$ in one $\mb\Theta_j$ with different scale. Thus by adjusting scale of diagonal blocks in different matrix $\mb\Theta_j$, we can make different network have different node parcelation.
To  simulate  the  connectivity  of variables among diagonal and off-diagonal blocks, we control the connectivity of each variables on diagonal block with a high density, then giving a low density to each off-diagonal block. Following the above steps, we generate several different $\mb\Theta_j$ and $\mb{H}_j$. 
Then each $\mb\Theta^\star_j$ can be derived from $\mb{H}_j^\top \mb\Theta_j \mb{H}_j$.

Given $\mb\Theta_j$, we can thus obtain $\mb{\Sigma}_j$, which is the inverse of $\mb\Theta_j$.
Due to the assumption of the independence of each Gaussian distribution, we obtain the covariance matrix $\mb{\Sigma}$ of the mixture Gaussian distribution. Then we generate $n$ samples randomly from the mixture Gaussian distribution. 

\subsubsection{Compared methods}
To demonstrate the effectiveness of our proposed method, we test against several state-of-art methods coherent brain network discovery methods:
\begin{itemize}
    \item \textit{CGL} \cite{yin2018coherent}: CGL aims to achieve node discovery and directed edge detection at the same time. Meanwhile, it can distinguish direct links from indirect connections due to its solid probabilistic formulation. 
    \item \textit{ONMtF} \cite{bai2017unsupervised}: ONMtF also aims to complete node discovery and edge detection at the same time. However, it focuses on explaining the spatial continuity of results. 
    We only apply it on the task of nodes discovery, due to its inability of directed edge discovery.
    \item \textit{\emph{k}-means + CGL}: 
    This pipeline method is more appropriate than CGL for the problem defined in this paper. We first employ \emph{k}-means to assign each $\bsym{x}_i$ to different nodes, then using CGLasso for each group to obtain the final ${\hat{\mb{\Theta}}}^\star_j$ and $\hat{\mb{H}}_j$.
    \item \textit{\emph{k}-means + OMNtF}: 
    This is also a pipeline method that first splits the whole sample of $\bsym{x}_i$ into different nodes by using \emph{k}-means, then using ONMtF on each node to obtain each ${\hat{\mb{\Theta}}}^\star_j$ and $\hat{\mb{H}}_j$.
    \item \textit{\emph{k}-means + JGL} \cite{gao2016estimation}: A Joint Graphical Model is proposed in \cite{gao2016estimation}, which aims to discover a mixture Gaussian distribution. However, it applies to the level of nodes. We therefore  employ \emph{k}-means to map $\bsym{x}_i$ into the node space of $\bsym{y}_i$ first.
\end{itemize}
\subsubsection{Experiment Setting}
We simulate four scenarios by changing one parameter and keeping the others fixed. Each scenario aims to study one of aforementioned research questions (\textit{RQ}). In these situations, we select sample size $n$, the standard error of noise $\mb{\sigma}$, the variables number $p$ of $\bsym{x}_i$, and the group number $k$ as the controlled parameters. 
\begin{itemize}
    \item \textit{Scenario 1}: We fix $p=70$ (the number of variables), $\mb{\sigma}=0$ (the standard error of noise), $k=5$ (the number of nodes) and then control sample size $n$ from 200 to 2000. 
    \item \textit{Scenario 2}: We fix $n=2000$, $p=70$ and $k=5$, meanwhile control $\mb{\sigma}$ from 2 to 5. 
    \item \textit{Scenario 3}: We fix $n=2000$, $\mb{\sigma}=0$ and $k=5$, and then control $p$ from 70 to 350. 
    \item \textit{Scenario 4}: We fix $n=2000$, $\mb{\sigma}=0$, and then control $k$ from 3 to 11. 
\end{itemize}
To generalize the results of comparative experiments, we sample 10 times for all experiments and average their results to evaluate the precision and stability of our model. 
\subsubsection{Evaluation Protocol}

To evaluate the quality of edge detection, we employ Accuracy and F1-score in the comparative experiments. We follow \cite{yang2015fused} to define the accuracy and F1-score of edge detection: 
\begin{align}
    &Accuracy = \frac{n_d}{n_g},\\
    &F1 = \frac{2n_d^2}{n_an_d + n_gn_d},
\end{align}
where $n_d$ is the number of true edges detected by the algorithm, $n_g$ is the number true edges and $n_a$ is the total number of edges detected. Higher accuracy score or higher F1 score indicates better quality of edge detection. To evaluate the quality of clustering, we follow \cite{sun2009rankclus} to use the purity score  and normalized mutual information score (NMI). Higher purity score or higher NMI score indicates better quality of clustering.
\subsection{Comparative Results}
\input{rq1.tex}

To study the effect of sample size on the performance of MNGL, we design comparative experiments based on \textit{Scenario 1}.
Figure \ref{fig:rq1} shows the comparative results. 
We compare our proposed model with five baseline methods. 
The first row shows the results of the comparison on edge detection; the second row shows the results for node discovery. 
In the results of all scenarios, we use the same symbol, which is illustrated in the caption below Figure \ref{fig:rq1}. 
From the results in the first row of Figure \ref{fig:rq1}, we observe that the sample size $n$ indeed affects some methods, especially ONMtF and its derivations. As the sample size increases, the accuracy of these two methods become much higher. Encouragingly, this factor has no significant effect on our model. 
Overall, we can clearly see that our method MNGL is more accurate and robust than other methods as the sample size $n$ changes. Meanwhile, $n$ does not have a significant impact on the performance of MNGL, which means that our model performs well even with a small training set.

\input{rq2.tex}

To study the effect of noise on the performance of MNGL, we use the experimental setup \textit{Scenario 2}.
Our results are shown in Figure \ref{fig:rq2} where the horizontal axis in the figure represents the standard error of noise $\mb{\sigma}$. 
The larger the $\mb{\sigma}$, the stronger the noise.
It leads to smaller signal-to-noise ratio, which means it is more difficult to mine the network structure from the available samples.
As seen in the four sub-graphs, 
we find that noise affects all compared methods.
In particular, while JGL suffers the most influence, ONMtF and derivatives of it are more robust than CGL and its derivatives in this scenario. Meanwhile, our method, MNGL, is better than all other comparison methods in this experiment for both edge detection and node discovery. Furthermore, $\mb{\sigma}$ does not significantly decay the performance of MNGL.

\input{rq3.tex}

To study the effect of the number of variables on the performance of MNGL, we next use \textit{Scenario 3}, the results for which are shown in Figure \ref{fig:rq3}.
For the node discovery task, we see that the dimension of feature space has no impact on the performance of any methods. However, for edge detection, when the dimension is low, the performance of CGL and its derivatives outperforms the other baseline methods. In particular, as the dimension increases, the accuracy of CGL and its derivatives shows a significant downward trend compared to the others. Again, as expected, in this scenario the performance of MNGL is more accurate and robust than the baseline methods.

\input{rq4.tex}

To study the effect of the number of states on the performance of MNGL, we turn to \textit{Scenario 4} and report our findings in Figure \ref{fig:rq4}. 
Across all sub-figures, as the number of nodes increases, the robustness of all compared methods shows a downward trend. 
Specifically, for the edge detection task, the accuracy of CGL and its derivatives perform slightly better than ONMtF and its derivatives.
However, when considering the F1-Score, ONMtF outperforms the CGL methods.
For node discovery, ONMtF and its derivatives show more robustness than other baseline methods.
Overall, MNGL still significantly outperforms the compared methods in these settings, though there is a small degree of fluctuation.

Combining the four \textit{RQ}s raised above and the results of all these comparative experiments, we can draw the following four conclusions: First, ONMtF and its derivatives are not as good as other methods in the case of insufficient samples.
Second, CGL and its derivatives are more restrictive in high-dimensional space.
Third, Both the accuracy and robustness of all comparative methods will decrease under the impact of noise and the number of nodes.
Fourth, compared to the alternative methods, our proposed method MNGL exhibits greater accuracy and robustness in each scenario, indicating that neither sample size $n$, the dimension of feature space $p$, noise $\mb{\sigma}$ nor the number of nodes $k$ significantly degrades the performance of our method.

\subsection{Real-World Datasets}

We also evaluate our proposed method on the fMRI dataset from the ADHD-200 project
. Attention Deficit Hyperactivity Disorder, or ADHD, is a chronic and sometimes-devastating condition affecting 5-10\% of school-age children.
It is also extremely costly to treat -- the United States alone has spent more than 36 billion on ADHD \cite{adler2015diagnosing}. 
This real-world dataset is distributed by nilearn
. Specifically, there are 40 subjects in total. Among them, 20 subjects are labeled as ADHD, and the others are labeled as typically developing children (TDC). The fMRI scan of each subject in the dataset is a series of snapshots of 3D brain images of size $91 \times 109 \times 91$ over $\sim$176 time steps. Because the fMRI scanning datasets are contain only voxels, the nodes and connectivity among them are all unknown. 
\cite{bai2017unsupervised,yin2018coherent} put these two tasks in a unified model to find the optimal solution. However, they ignore the assumption of mixture network structure we defined in this paper. 
Furthermore, this part of the experiment lacks ground-truth as a reference to measure the accuracy and robustness of the model. Therefore we must consider the interpretability and rationality of the results. Specific to our proposed model, we are primarily concerned with whether or not our model can mine different cognitive networks from the fMRI datasets (various node assignments and functional connectivity).
Therefore for this subsection, we focus solely on applying our proposed method, MNGL, to this challenging task.

In our experimental setting, we focus on the multi-state brain network discovery among the same subjects, and report the results of both nodes discovery and edge detection.
To assign the voxels that can be considered as parts of the brain, we use anatomical automatic labeling (AAL) brain-shaped mask, which is provided by neurology professionals. We follow \cite{kuo2015unified} and use a middle slice of these scan for the ease of presentation. Consequently, each of the brain scans can be represented by about 3281 voxels. So it is more conventional for the visualization of the results. The datasets is a 3281 (variables) $\times$ 2992 (time steps) datasets and reasonably assume that they are drawn from a mixture Gaussian distribution. However, the number of Gaussian distributions $m$ and the number of nodes $k$ are both unknown and need to be selected in advance. 
Through repeated experimental observations, we find that $m=2$ and $k=6$ can provide the most reasonable results on the data sets.

\input{fmri_1.tex}

Figure \ref{fig:brain2} 
shows the multi-state network discovered by MNGL on the fMRI datasets. 
The results of edge detection and node discovery are shown on the first and second line, which corresponds to the functional network of state $\mathcal{S}_1$ and $\mathcal{S}_2$ respectively.
Meanwhile, each inferred node is displayed on the third and fourth line individually. 
First, we can see the difference between the two networks from the discovered edges and nodes.
We deliberately mark the differences between the nodes of two networks with red circles. More specifically, in the first line of Figure \ref{fig:brain2}, we find a strong and complete default mode network (DMN) for ADHD subjects, corresponding to group 3 and 5 in the third line.
A DMN is a network of interacting brain regions known to have activity highly
correlated with each other while being distinct from other networks in the brain,
including the Parietal and Occipital Lobes, the Cingulum Region Posterior, and the Frontal Cortex.
However, in the second line of Figure \ref{fig:brain2}, this mode is not intact.
In particular, the Frontal Cortex is missing in the network, while the rest of the connections are different from the functional network in the first line. The specific relationship between each node can be found in the sub-figures of discovered edges.

\input{tdc_1.tex}
For comparison, we apply MNGL again on the subjects of TDC, which represent the group of typically-developing children. We can observe from Figure \ref{fig:tdc1} that, although there are differences in the node assignments and the connectivity structure among each node, there is no deletion of DMN in each network. At the same time, the network of state $\mathcal{S}_1$ from TDC and that from ADHD have a certain degree of similarity from the nodes result to the connectivity structure, which allows us to have more reference when analyzing the differences and connections between the two subjects. 
These results give us reason to believe that the brain scans of these subjects have some similar functional structure correspondences similar to on-task and off-task states.

Despite the lack of ground-truth, we believe that the current results are still consistent with the problem defined in this paper: We find strong evidence that there is a multi-state brain cognitive network in the fMRI datasets, and our proposed model MNGL can effectively mine this mixture network structure. 

\section{Related Works}

Existing works can be divided into two categories.
Firstly, for coherent brain network discovery, 
ONMtF~\cite{brunet2004metagenes} is a useful pattern recognition method. \cite{bai2017unsupervised} extend ONMtF by adding a spatial continuity penalty, which can increase the interpretability of the parcellated regions. This method is a coherent model which can output the result of nodes discovery and edge detection simultaneously. However, it has discovered the edges based on the correlation matrix instead of inferring direct links between each node. 
Instead of using a correlation matrix, \cite{friedman2008sparse,banerjee2008model} focus on sparse inverse covariance estimation for discovering connectivity of brain network based on large-scale datasets. These kinds of methods can distinguish direct links from indirect connections due to their solid probabilistic foundation.
\cite{yin2018coherent} propose a model called CGL to achieve the coherent brain network discovery, including edge detection and node discovery. However, this method ignores the problem of multi-state problem we mentioned in this paper.

For the multi-state problem, we consider the Gaussian Mixture Model (GMM) \cite{pearson1894contributions}. GMMs model the distribution of data observations as a weighted sum of parameterized Gaussian distributions. 
However, a prominent issue related to GMM is estimating the parameters given observations \cite{back1993implied}. 
Through many extensions, the EM algorithm has proven to be a powerful algorithm for the maximum-likelihood estimation of GMMs~\cite{dempster1977maximum}. 
Additionally, \cite{schwarz1978estimating,akaike1974new} consider the issue of the number of mixture components in the model, which can lead to over-fitting in practice. 
GMMs have been widely used in many areas, especially for network discovery \cite{newman2007mixture,zhu2014structural,mclachlan2000finite}. 
Most existing studies for mixture modeling focus on regularizing only the mean parameters with diagonal covariance matrices \cite{wang2008variable, xie2008penalized}, though some works \cite{zhou2009penalized, hill2013network, wu2013cancer} have started considering regularization of the covariance parameters.
However, these works do not touch on the key issue of identifying the varying sparse structures of the precision matrices across the components of a mixture model in brain network discovery.
\cite{gao2016estimation} proposes a joint graphical model (JGL) to deal with cluster-specific networks.
\cite{yin2020gaussian} aims to edge detection task by combing graphical lasso with GMM. However, these models need brain parcellation to be given first. 
Thus, existing models related to GMM are thus not suitable for the special problem defined in this paper.



\section{Conclusion}

In this work, we define the open problem of multi-state brain network discovery, which is to infer various brain parcellations and connectivities across different brain states.
Previous works on brain network discovery derive an average brain network based on the assumption that only one single activity state of the brain generates the signals.
However, according to recent studies in the area of brain network, assuming single-state networks ignores a crucial of cognitive brain networks.
To better understand the temporally-changing functional network of the brain, we propose a novel model called MNGL, which can discover \textit{multiple} brain networks, including nodes and their connectivity based on on only unlabeled fMRI scans. 
Through extensive controlled experiments, we demonstrate that our proposed model shows more effectiveness and robustness than other baseline models.
MNGL also shows expected and meaningful results on the real ADHD-200 fMRI dataset. 
We thus have reason to believe that our method can be applied in multi-state brain network for a better understanding of brain function and behavioral performance.

\section{Acknowledgments}
\label{sec:ack}
Hang Yin and Xiangnan Kong was supported in part by NSF grant IIS-1718310. Yanhua Li was supported in part by NSF grants IIS-1942680 (CAREER), CNS-1952085, and DGE-2021871.

\footnotesize{
\balance
\bibliographystyle{unsrtnat}
\bibliography{header,reference}}
\end{document}

%% file: header.tex
\usepackage{amsthm}
\usepackage{algorithm}
\usepackage{algpseudocode}
\usepackage{amsmath}

\usepackage{amssymb}
\usepackage{balance}
\usepackage{booktabs}
\usepackage{bm}
\usepackage{color}
\usepackage{natbib}
\setcitestyle{compress,unsrtnat,comma,numbers,square}
\usepackage{multirow}
\usepackage{stmaryrd}
\usepackage{subcaption}
\usepackage{url}
\usepackage{thmtools}
\usepackage{ctable}
\declaretheoremstyle[
  spaceabove=6pt, spacebelow=6pt,
  headfont=\scshape,
  notefont=\itshape, notebraces={(}{)},
  bodyfont=\itshape,
  postheadspace=0.5em,
  numberwithin=section,
  headpunct=:
]{mystyle}

\newcommand{\eg}[0]{\textit{e.g.},\ }

\newcommand{\ie}[0]{\textit{i.e.},\ }  
\newcommand{\bsym}{\boldsymbol}
\newcommand{\mb}{\mathbf}
\newcommand{\mc}{\mathcal}

%% file: butterfly.tex
\begin{figure}[t]
\centering
\begin{minipage}{\columnwidth}
    \includegraphics[width=1\textwidth]{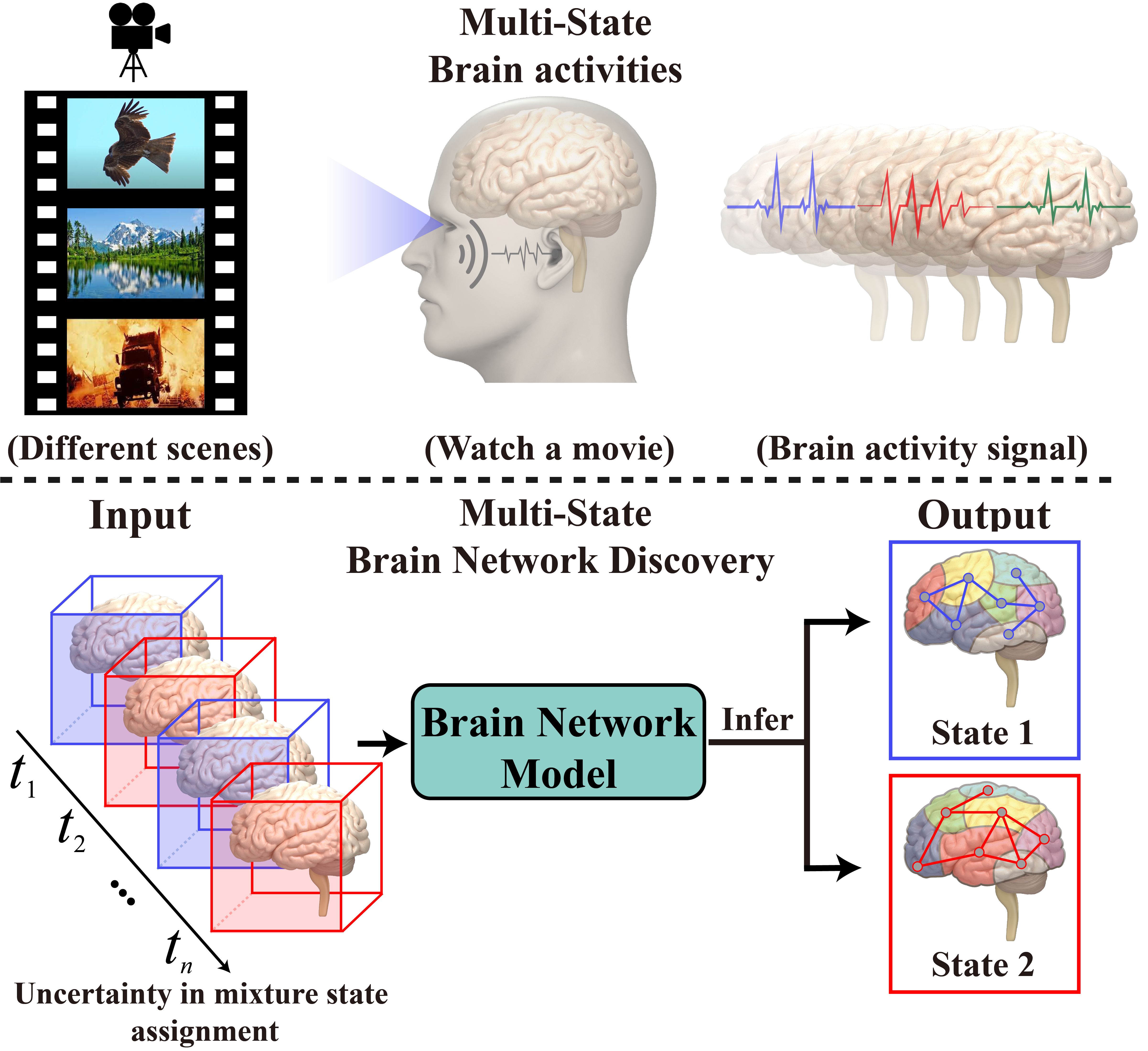}
\end{minipage}
\vspace{-2pt}
\caption{The problem of multi-state brain network discovery. Brain activities over time may derive from the mixture of multiple brain states (\eg different brain states appear during different scenes of a movie). Without knowledge of mixture state assignment, our goal is to discover the multiple underlying brain network states, allowing for differing brain parcellation and connectivity.}
\label{fig:butterfly}
\vspace{-20pt}
\end{figure}

%% file: family.tex
\begin{figure*}[t]
\centering
\begin{subfigure}{1\columnwidth}
\centering
  \includegraphics[width=.9\linewidth]{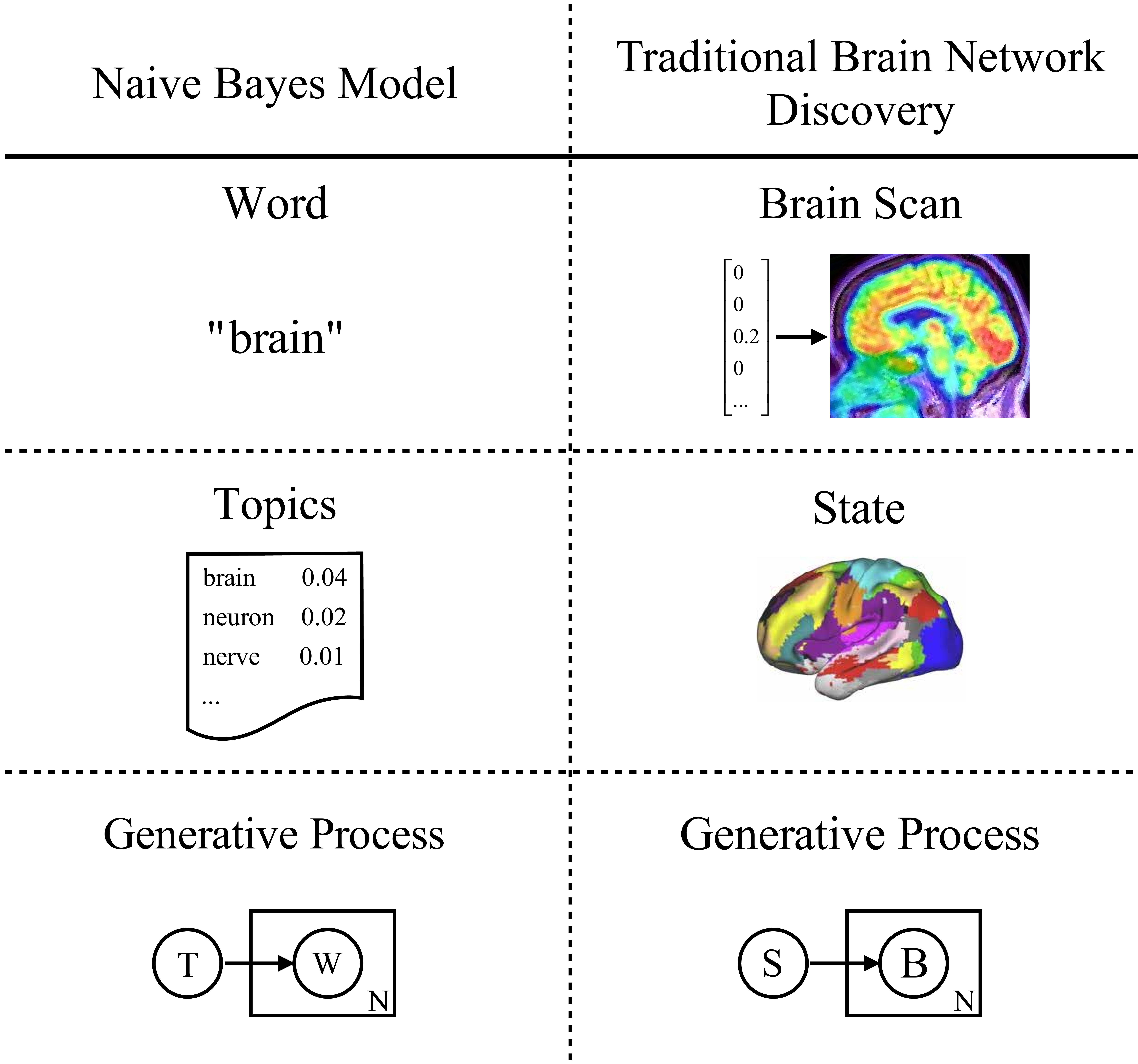}
  \caption{Naive Bayes~\cite{mccallum1998comparison} and Brain Network Discovery Model~\cite{liu2017collective,bai2017unsupervised}}
  \label{fig:fam1}
\end{subfigure}
\begin{subfigure}{1\columnwidth}
\centering
  \includegraphics[width=.9\linewidth]{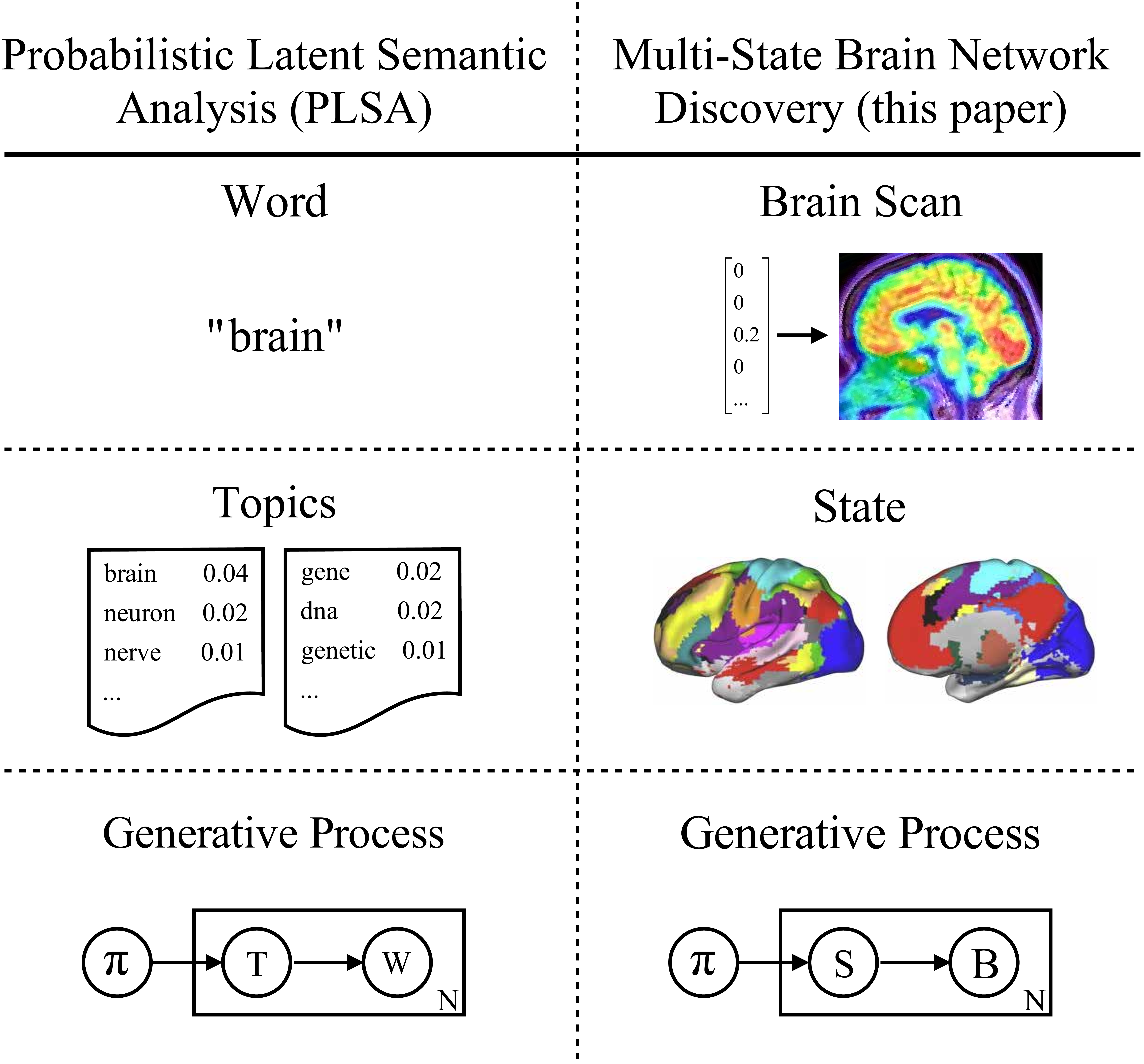}
  \caption{PLSA\cite{hofmann1999probabilistic} and Multi-State Brain Network Discovery Model (ours)}
  \label{fig:fam2}
\end{subfigure}

\caption{Two pairs of comparison: (a) naive bayes model and traditional brain network discovery model, (b) PLSA and our model in this paper. In each generative process, the boxes are "plates" representing replicates.
The outer plate represents document in naive bayes and PLSA, or observation subject in brain network study, while the inner plate represents the generative process of word ($\text{W}$) in a given document or brain scan ($\text{B}$) in a given subject, each of which word or brain scan is associated with a choice of topic ($\text{T}$) or state ($\text{S}$).
$\pi$ is the topic or state distribution. $\text{N}$ denotes the number of words or scans. 
}
\vspace{-15pt}
\label{fig:family}
\end{figure*}

%% file: notation.tex
\begin{table*}[t]
    \centering 
    \caption{Important Notations.}\vspace{0pt}\label{tab:notation}
    \begin{tabular}{|r|l|}
    \hline\vspace{-8pt}&\\
    Symbol& Definition\\
    \hline\vspace{-8pt}&\\
    $m$ & The number of gaussian distributions\\
    $\mb{X} \in \mathbb{R}^{n \times p}$ & $n$ observations of $p$-variate Gaussian distribution\\
    $\mb{H}_j \in \mathbb{R}^{p \times k}$ & The clustering indicator matrix and $0 \le j \le m$\\ 
    $\mb{Y} \in \mathbb{R}^{n \times k}$ & $n$ the projection of $\mb{X}$ along $\mb{H}$ matrix\\
    $\mb{\Sigma}_j \in \mathbb{R}^{p \times p}$ & The covariance of $p$-variate Gaussian distribution\\
    ${\mb{\Sigma}^\star}_j \in \mathbb{R}^{k \times k}$ & The projection of $\mb{\Sigma}$ along $\mb{H}$ matrix\\
    $\mb{\Theta}_j \in \mathbb{R}^{p \times p} $ & The true precision matrix of all variables\\
    ${\mb{\Theta}^{\star}}_j \in \mathbb{R}^{k \times k} $ & The true precision matrix of all nodes\\
    $\mb{\hat{\Theta}}_j \in \mathbb{R}^{p \times p} $ & The estimate of true precision matrix of all variables $\mb{\Theta}$\\
    $\mb{{\hat{\Theta}}^{\star}}_j \in \mathbb{R}^{k \times k} $ & The estimate of true precision matrix of all nodes $\mb{\Theta}^{\star}$\\
    $\phi_j$ & The prior probability of the $j$-th base distribution chosen to generate a sample\\
    $\gamma_{ij}$ & The posterior probability of the $i$-th observation generated by the $j$-th distribution\\
    
        \hline
    \end{tabular}
    \vspace{-15pt}
\end{table*}

%% file: rq1.tex
\begin{figure}[t]
\centering

\begin{subfigure}{0.49\columnwidth}
\centering
  \includegraphics[width=1\linewidth]{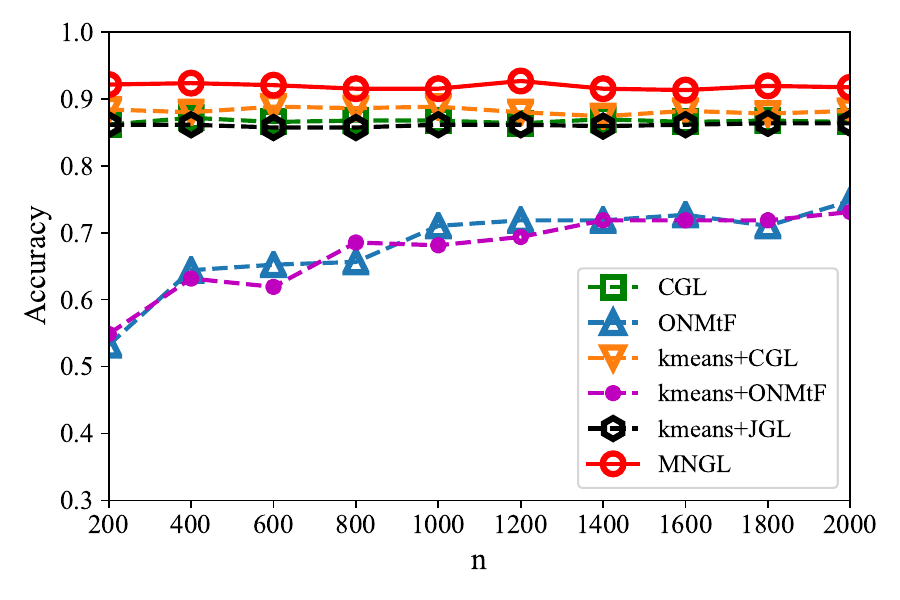}
  \caption{Accuracy of Scenario 1}
\end{subfigure}
\begin{subfigure}{0.49\columnwidth}
\centering
  \includegraphics[width=1\linewidth]{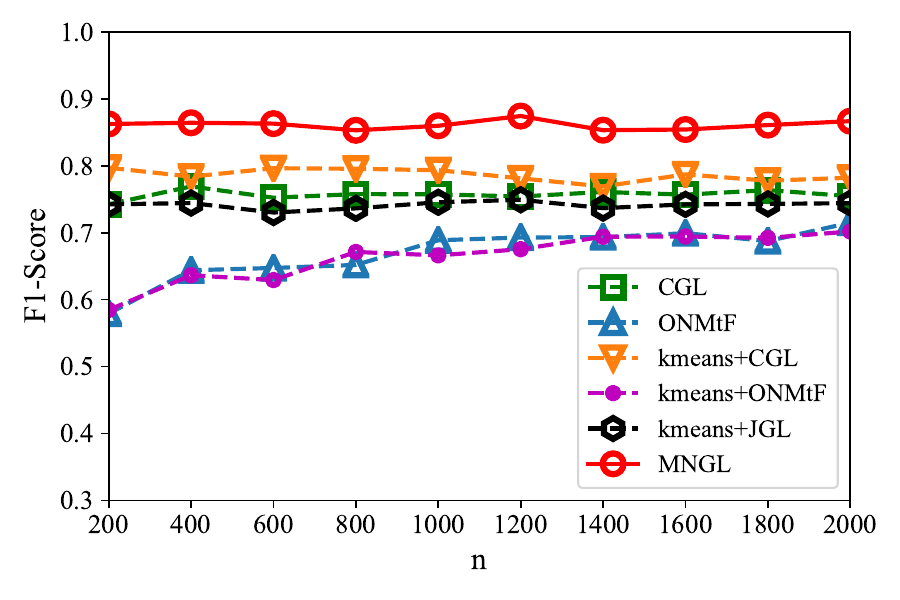}
  \caption{F1-Score of Scenario 1}
\end{subfigure}
\begin{subfigure}{0.49\columnwidth}
\centering
  \includegraphics[width=1\linewidth]{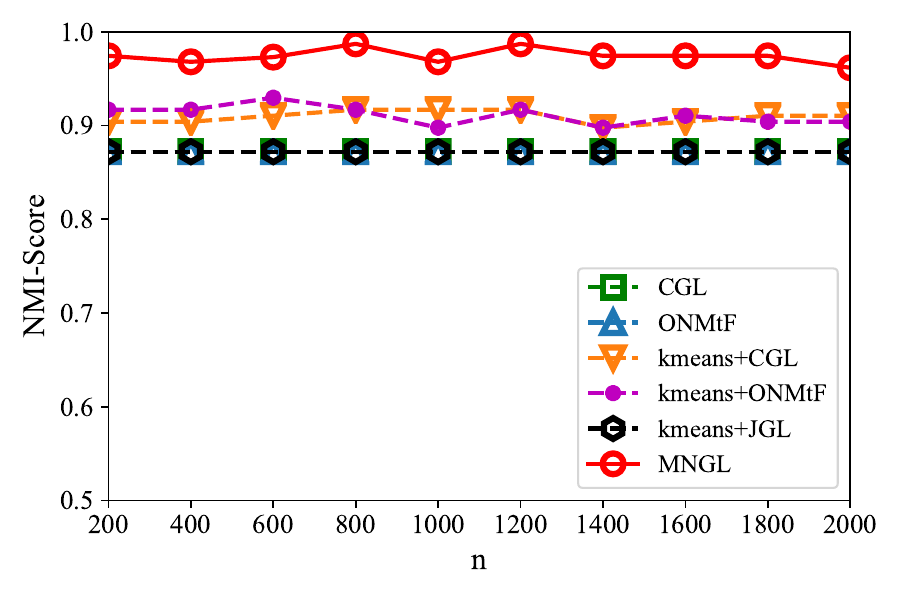}
  \caption{NMI of Scenario 1}
\end{subfigure}
\begin{subfigure}{0.49\columnwidth}
\centering
  \includegraphics[width=1\linewidth]{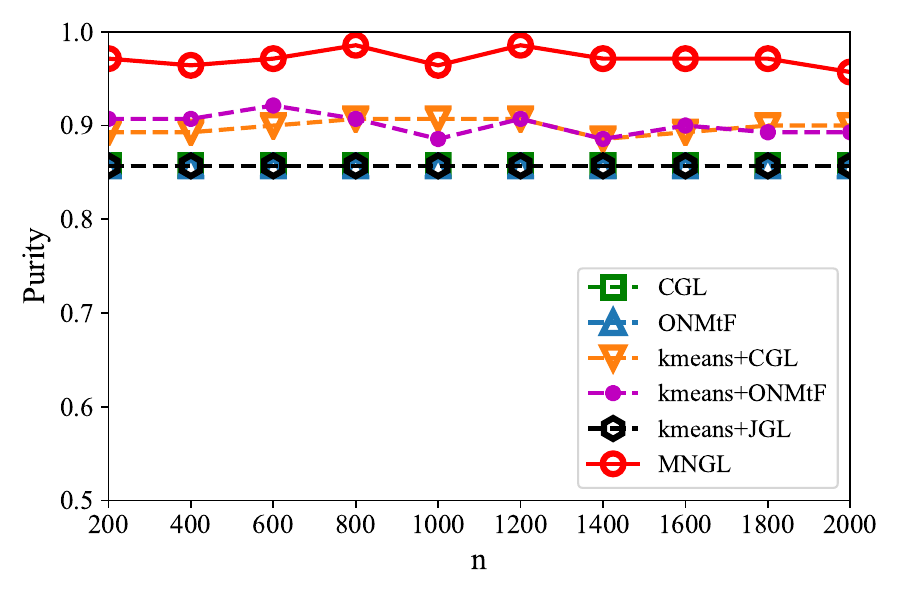}
  \caption{Purity of Scenario 1}
\end{subfigure}
\caption{Comparison of each method on edge detection and node discovery. 
The first row shows the results of edge detection, and the second shows the results of node discovery. 
The four sub-figures above consider different sample size $n$ from 200 to 2000. The other parameters are left fixed. 
}
\label{fig:rq1}
\vspace{-15pt}
\end{figure}

%% file: rq2.tex
\begin{figure}[t]
\centering

\begin{subfigure}{0.49\columnwidth}
\centering
  \includegraphics[width=1\linewidth]{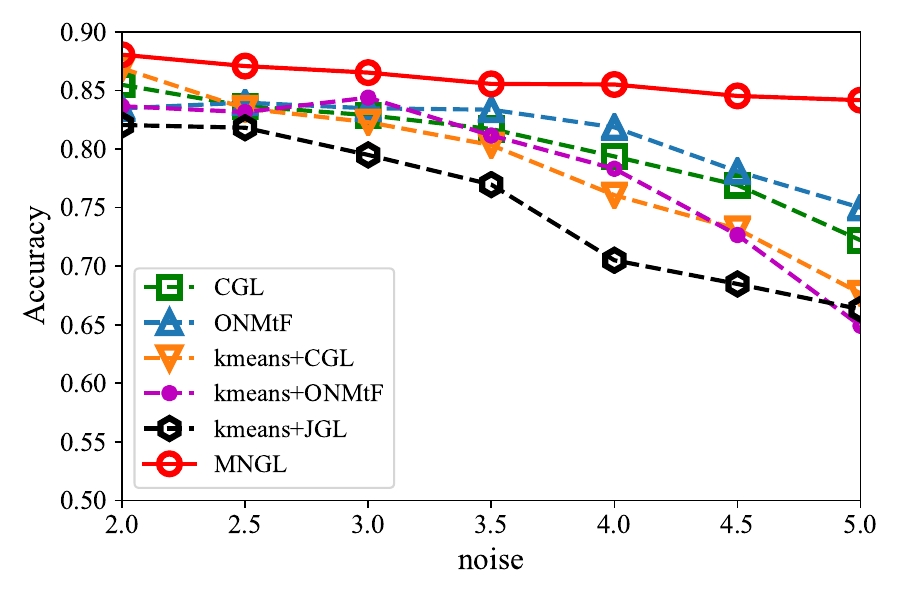}
  \caption{Accuracy of Scenario 2}
\end{subfigure}
\begin{subfigure}{0.49\columnwidth}
\centering
  \includegraphics[width=1\linewidth]{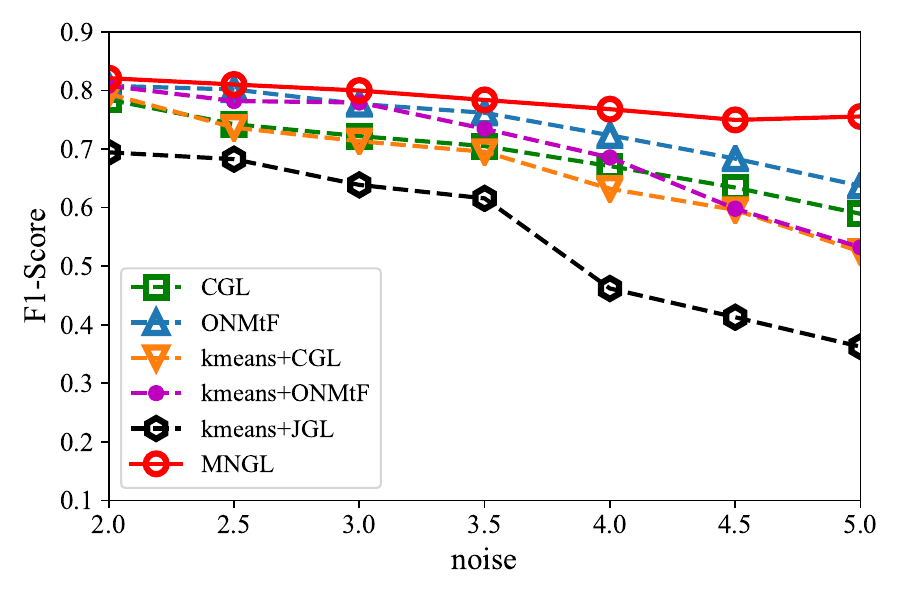}
  \caption{F1-Score of Scenario 2}
\end{subfigure}
\begin{subfigure}{0.49\columnwidth}
\centering
  \includegraphics[width=1\linewidth]{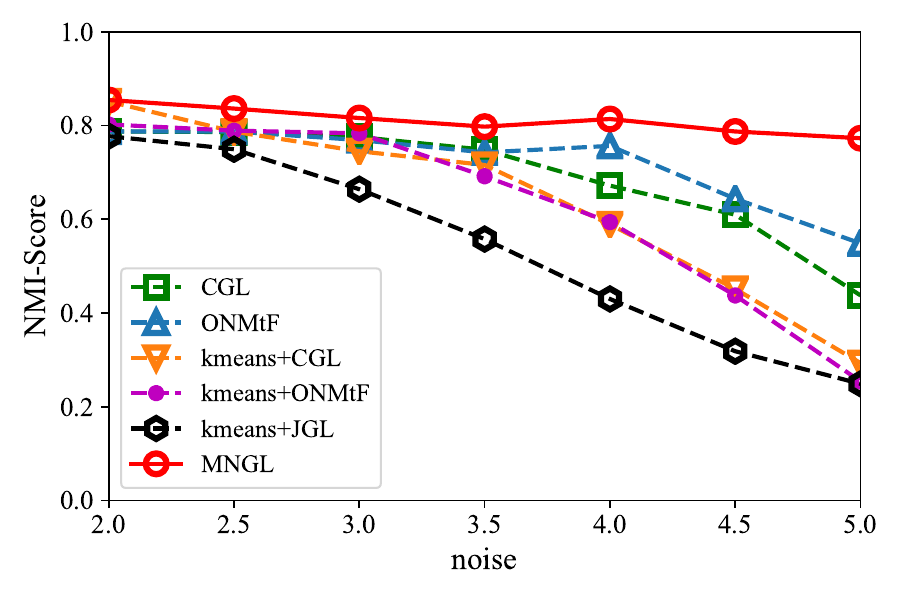}
  \caption{NMI of Scenario 2}
\end{subfigure}
\begin{subfigure}{0.49\columnwidth}
\centering
  \includegraphics[width=1\linewidth]{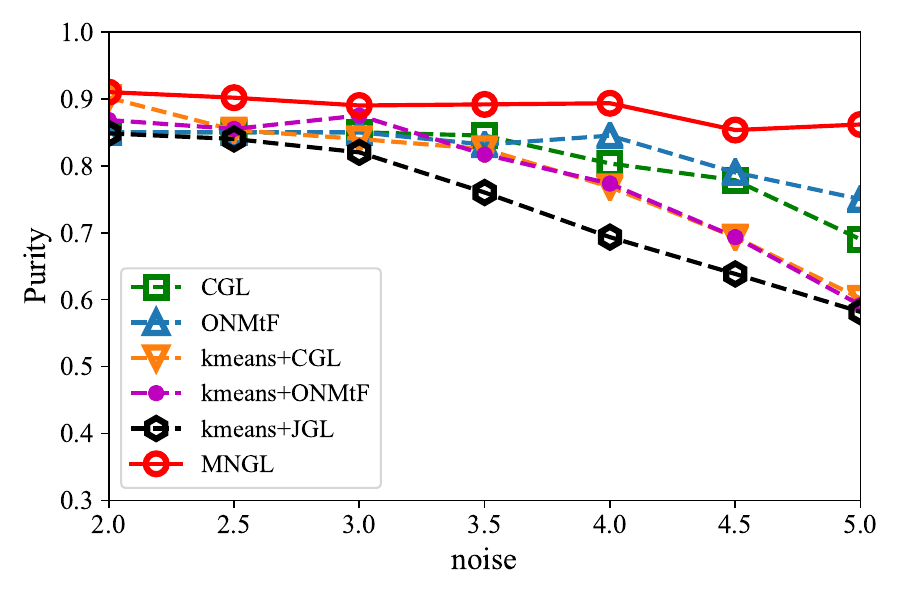}
  \caption{Purity of Scenario 2}
\end{subfigure}
\caption{The four sub-figures above consider different $\mb{\sigma}$ (the standard error of noise) from 2 to 5, meanwhile fix the other parameters, which correspond to scenario2;}
\label{fig:rq2}
\vspace{-15pt}
\end{figure}

%% file: rq3.tex
\begin{figure}[t]
\centering

\begin{subfigure}{0.49\columnwidth}
\centering
  \includegraphics[width=1\linewidth]{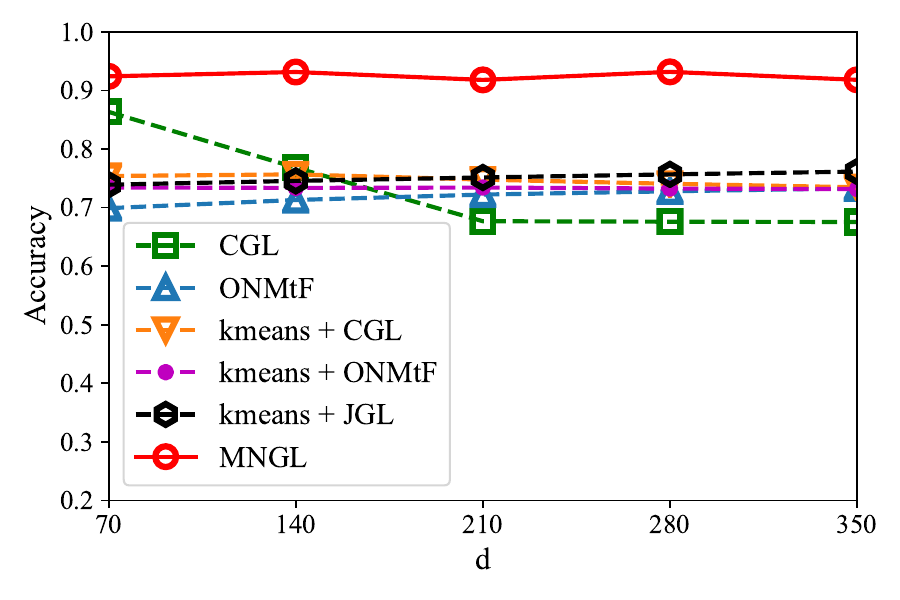}
  \caption{Accuracy of Scenario 3}
\end{subfigure}
\begin{subfigure}{0.49\columnwidth}
\centering
  \includegraphics[width=1\linewidth]{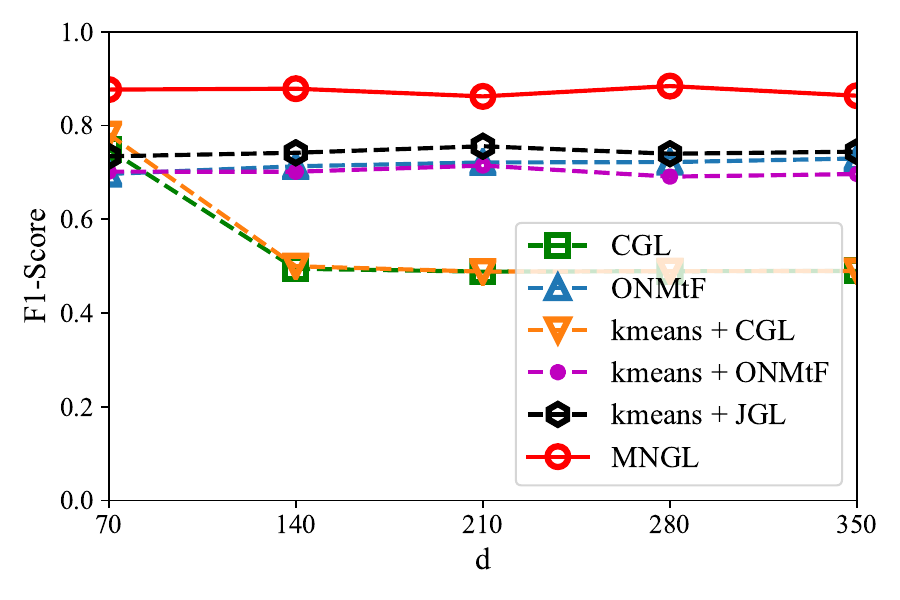}
  \caption{F1-Score of Scenario 3}
\end{subfigure}
\begin{subfigure}{0.49\columnwidth}
\centering
  \includegraphics[width=1\linewidth]{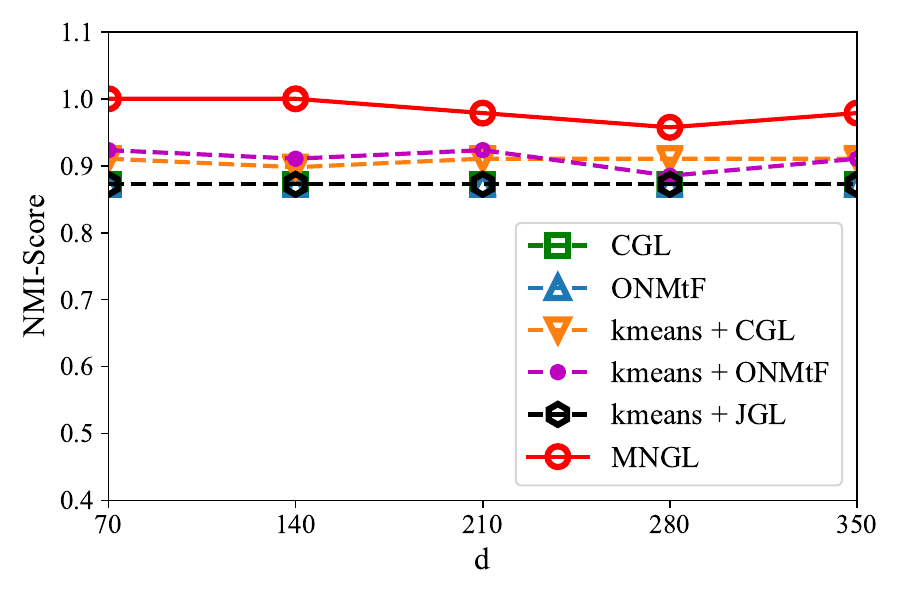}
  \caption{NMI of Scenario 3}
\end{subfigure}
\begin{subfigure}{0.49\columnwidth}
\centering
  \includegraphics[width=1\linewidth]{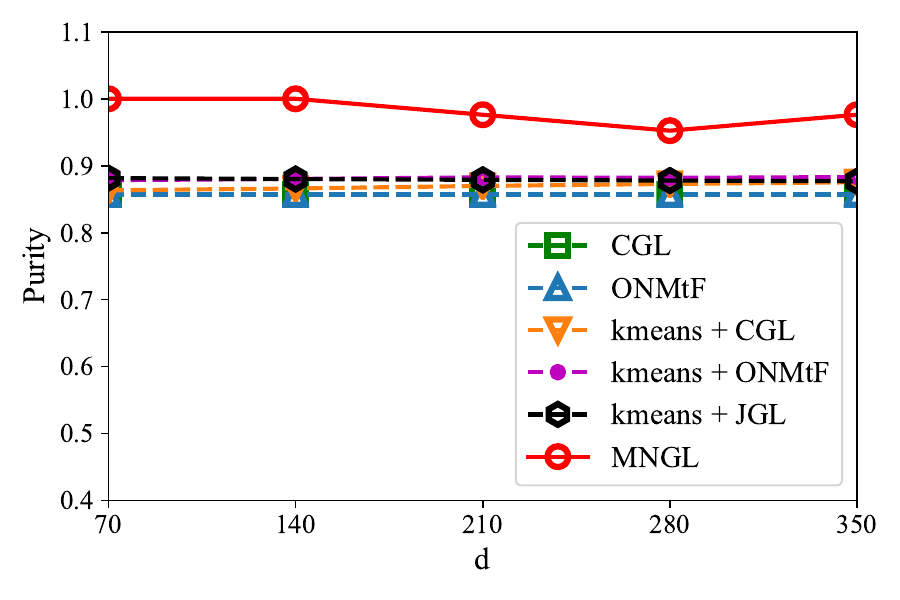}
  \caption{Purity of Scenario 3}
\end{subfigure}
\caption{The four sub-figures above consider different $p$ (the number of variables $\mb{x}_i$) from 70 to 350, meanwhile fix the other parameters, which correspond to scenario3;}
\label{fig:rq3}
\vspace{-15pt}
\end{figure}

%% file: rq4.tex
\begin{figure}[t]
\centering

\begin{subfigure}{0.49\columnwidth}
\centering
  \includegraphics[width=1\linewidth]{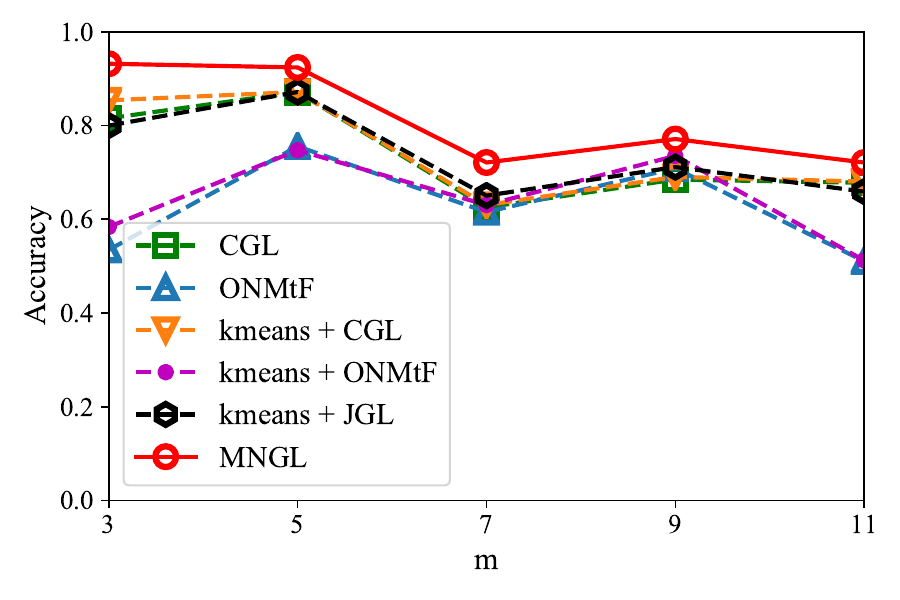}
  \caption{Accuracy of Scenario 4}
\end{subfigure}
\begin{subfigure}{0.49\columnwidth}
\centering
  \includegraphics[width=1\linewidth]{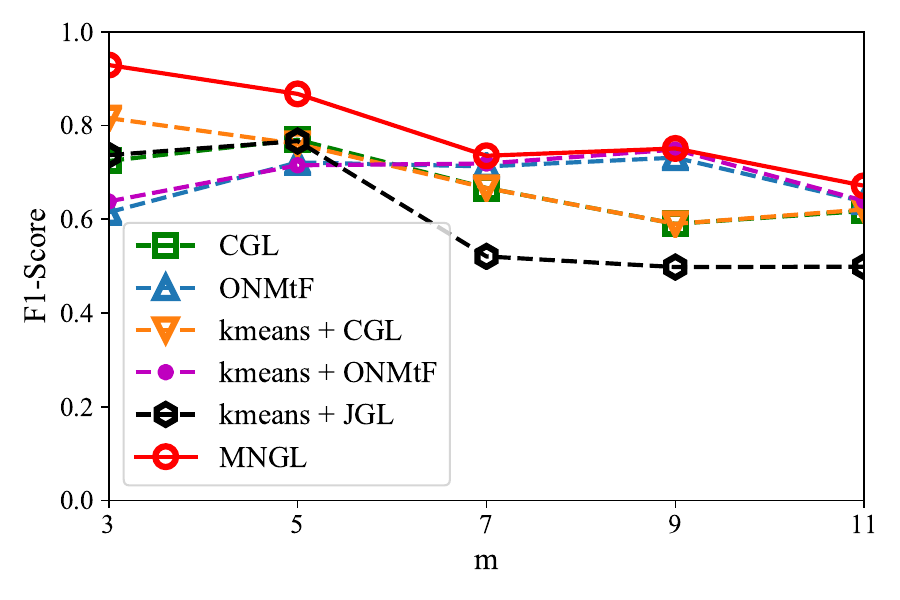}
  \caption{F1-Score of Scenario 4}
\end{subfigure}
\begin{subfigure}{0.49\columnwidth}
\centering
  \includegraphics[width=1\linewidth]{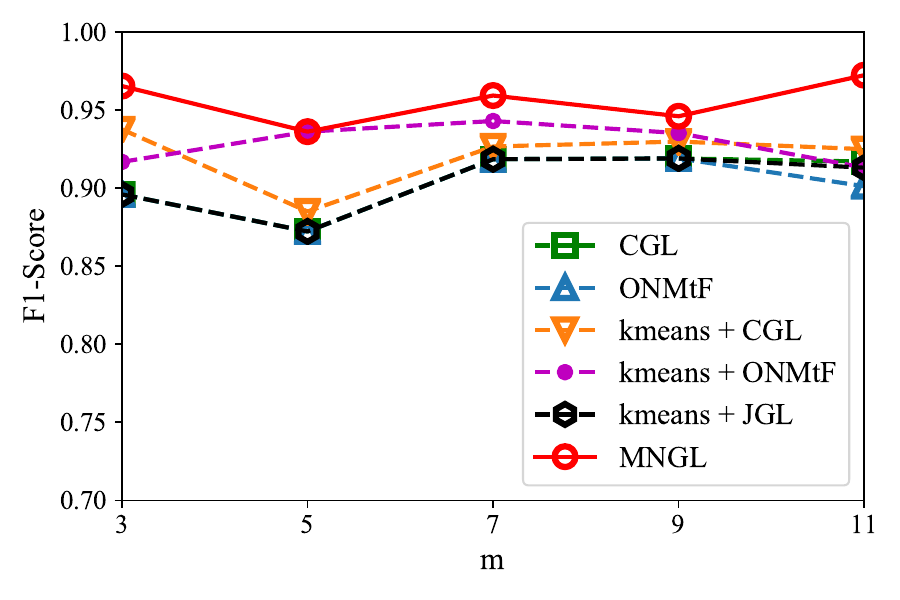}
  \caption{NMI of Scenario 4}
\end{subfigure}
\begin{subfigure}{0.49\columnwidth}
\centering
  \includegraphics[width=1\linewidth]{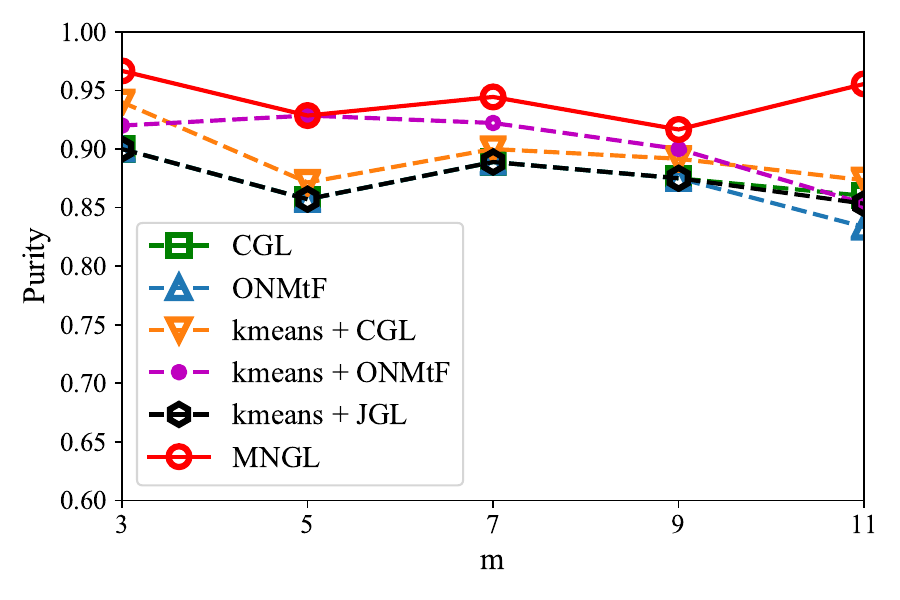}
  \caption{Purity of Scenario 4}
\end{subfigure}
\caption{The four sub-figures above consider different $k$ (the number of nodes $\mb{y}_i$) from 3 to 11, meanwhile fix the other parameters, which correspond to scenario4;}
\label{fig:rq4}
\vspace{-15pt}
\end{figure}

%% file: fmri_1.tex
\begin{figure}[t]
\centering
\begin{subfigure}{0.49\columnwidth}
\centering
  \includegraphics[width=.9\linewidth]{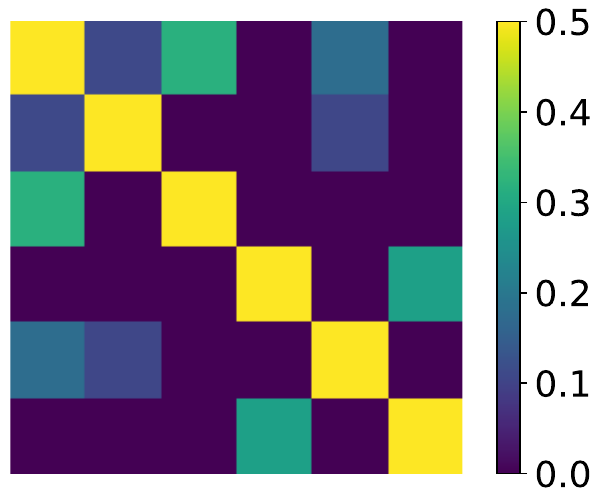}
  \caption{Edges of $\mathcal{S}_1$}
  \label{fig:edge1_adhd}
\end{subfigure}
\begin{subfigure}{0.49\columnwidth}
\centering
  \includegraphics[width=.9\linewidth]{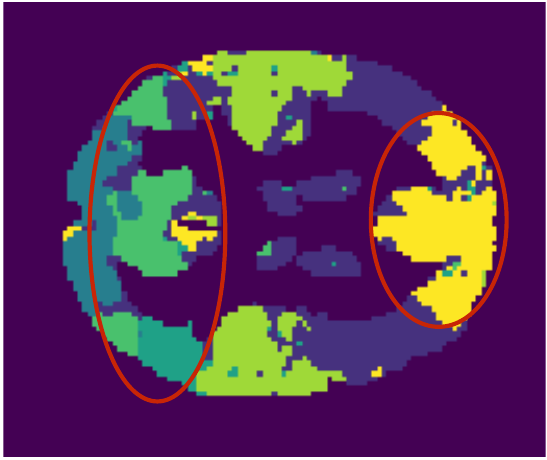}
  \caption{Nodes of $\mathcal{S}_1$}
  \label{fig:node1_adhd}
\end{subfigure}
\begin{subfigure}{0.49\columnwidth}
\centering
  \includegraphics[width=.9\linewidth]{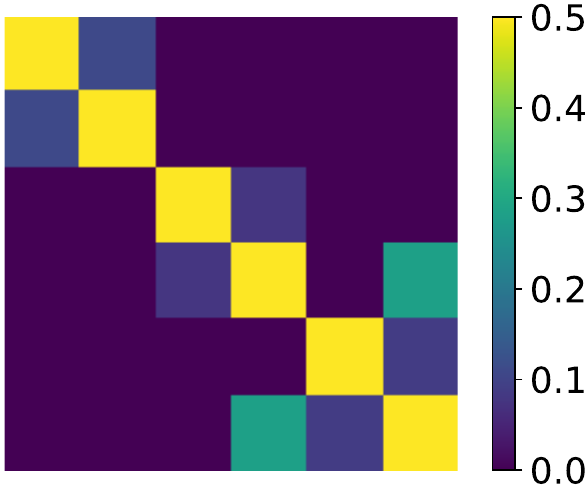}
  \caption{Edges of $\mathcal{S}_2$}
  \label{fig:edge2_adhd}
\end{subfigure}
\begin{subfigure}{0.49\columnwidth}
\centering
  \includegraphics[width=.9\linewidth]{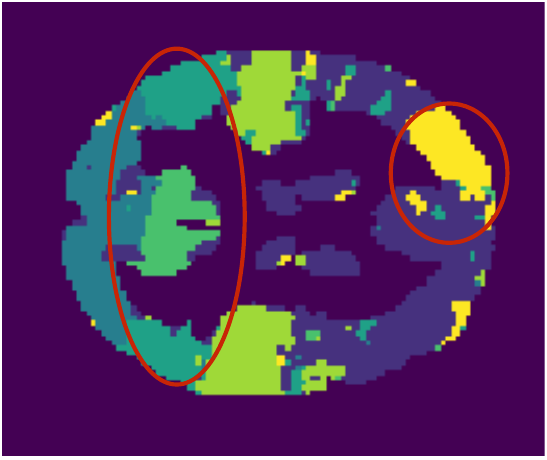}
  \caption{Nodes of $\mathcal{S}_2$}
  \label{fig:node2_adhd}
\end{subfigure}
\begin{subfigure}{1.\columnwidth}
\centering
  \includegraphics[width=0.15\linewidth]{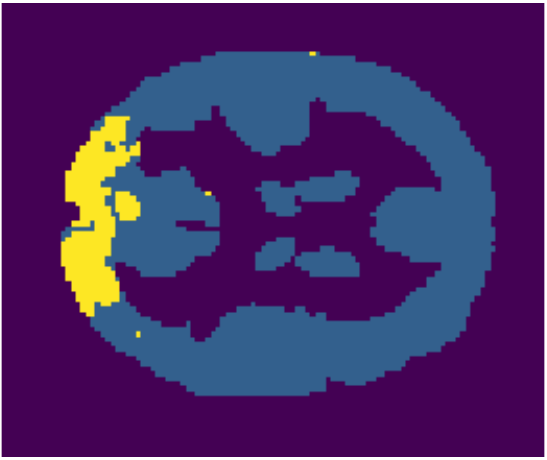}
  \includegraphics[width=0.15\linewidth]{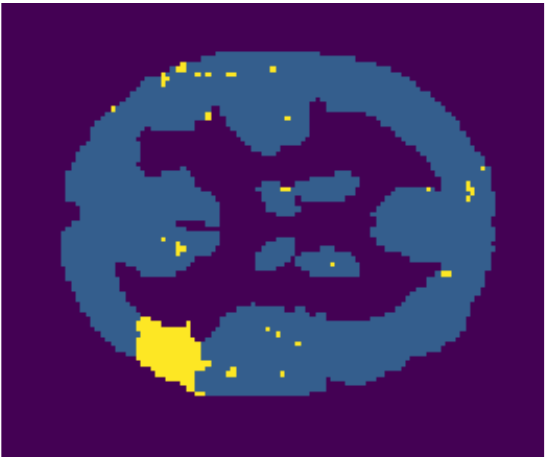}
  \includegraphics[width=0.15\linewidth]{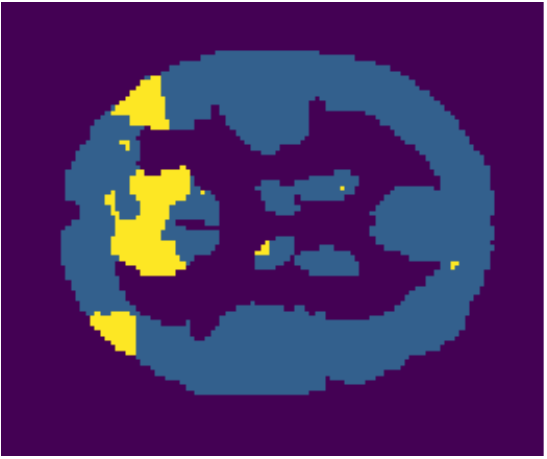}
  \includegraphics[width=0.15\linewidth]{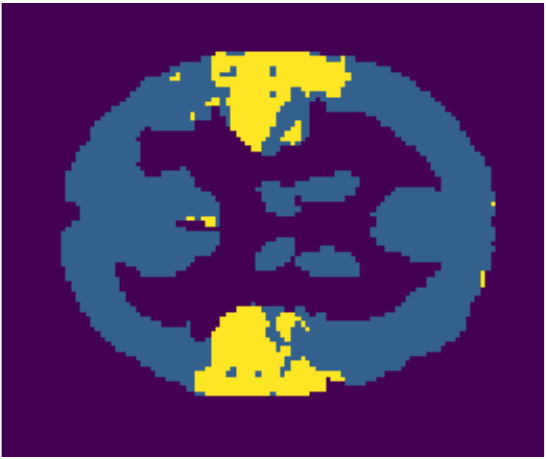}
  \includegraphics[width=0.15\linewidth]{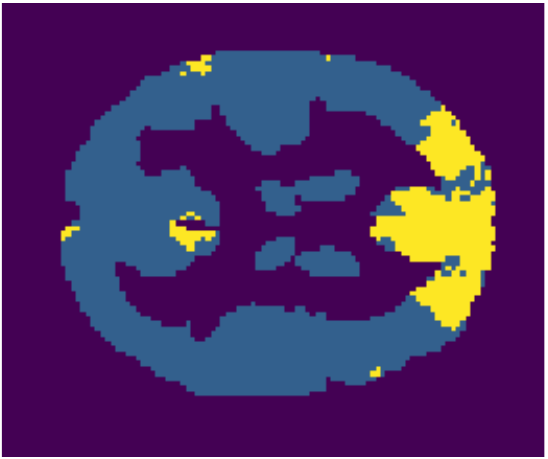}
  \includegraphics[width=0.15\linewidth]{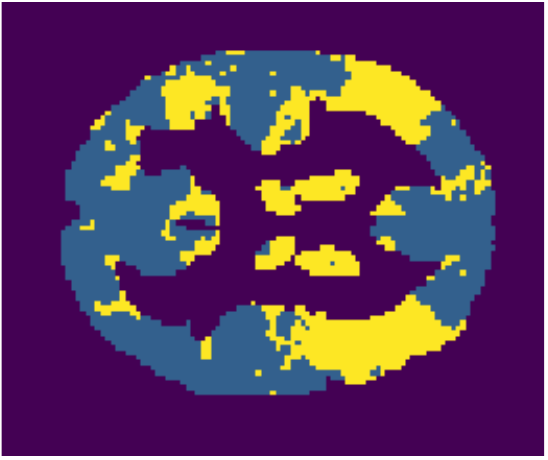}
  \caption{Nodes (1-6) of $\mathcal{S}_1$}
  \label{fig:ind1_adhd}
\end{subfigure}
\begin{subfigure}{1.\columnwidth}
\centering
  \includegraphics[width=0.15\linewidth]{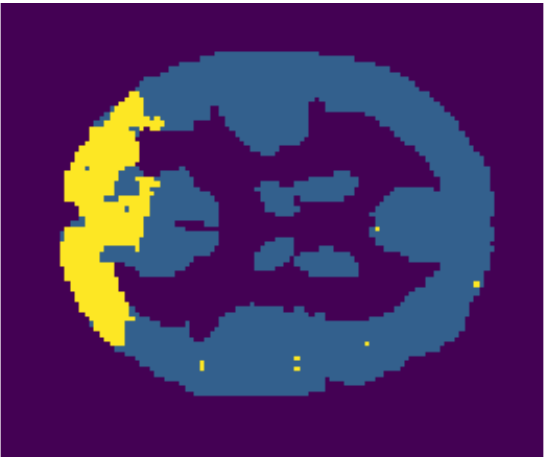}
  \includegraphics[width=0.15\linewidth]{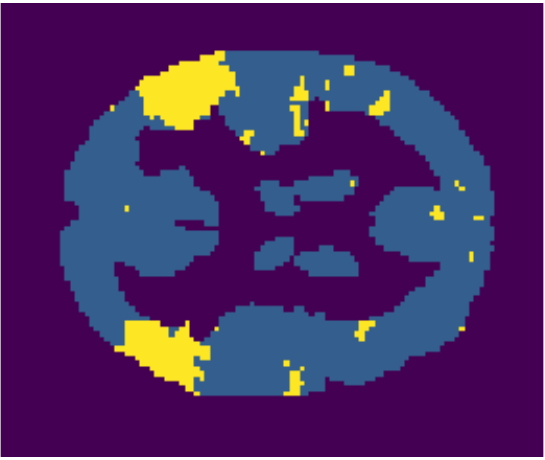}
  \includegraphics[width=0.15\linewidth]{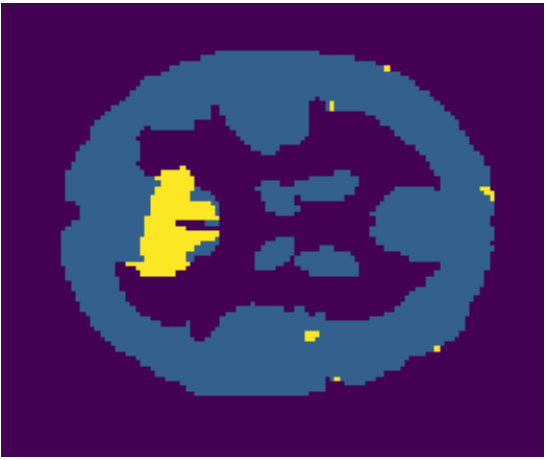}
  \includegraphics[width=0.15\linewidth]{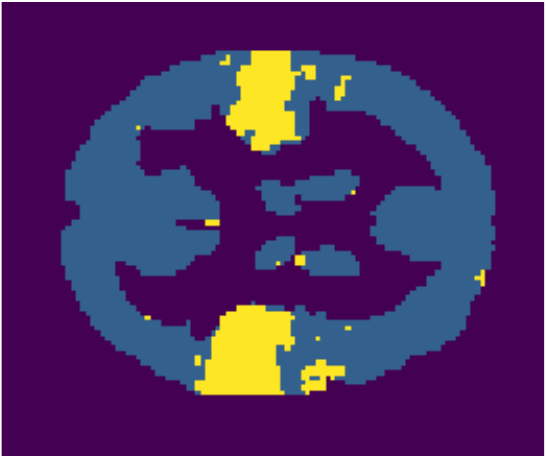}
  \includegraphics[width=0.15\linewidth]{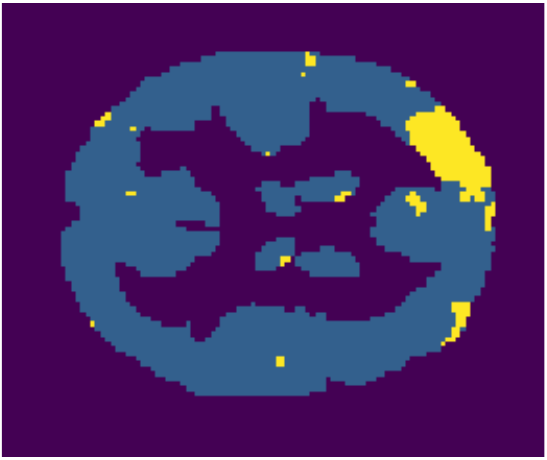}
  \includegraphics[width=0.15\linewidth]{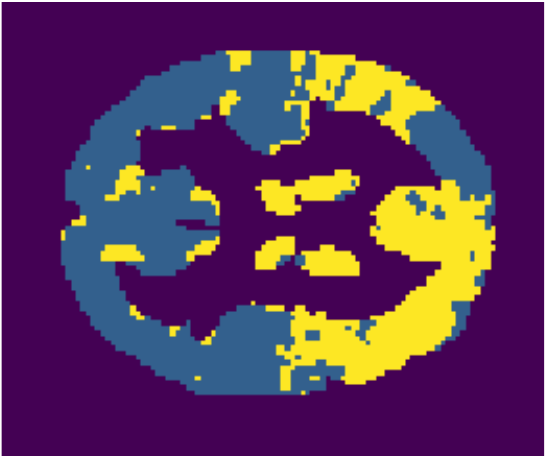}
  \caption{Nodes (1-6) of $\mathcal{S}_2$}
  \label{fig:ind2_adhd}
\end{subfigure}
\vspace{-15pt}

\caption{Discovered results of multi-state brain network in ADHD subjects ($k=6$). 
}
\label{fig:brain2}
\vspace{-15pt}
\end{figure}

%% file: tdc_1.tex
\begin{figure}[t]
\centering
\begin{subfigure}{0.49\columnwidth}
\centering
  \includegraphics[width=.9\linewidth]{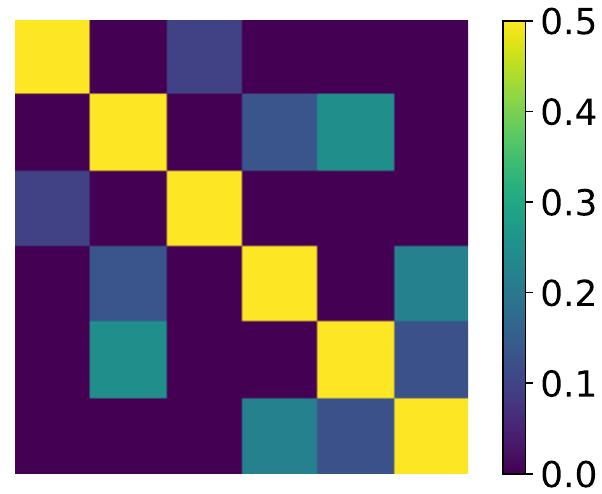}
  \caption{Edges of $\mathcal{S}_1$}
  \label{fig:edge1_tdc}
\end{subfigure}
\begin{subfigure}{0.49\columnwidth}
\centering
  \includegraphics[width=.9\linewidth]{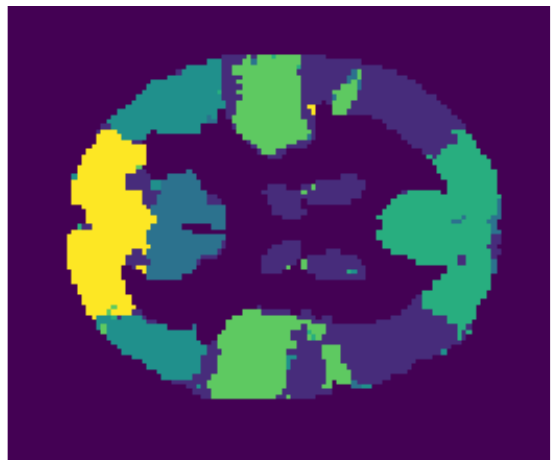}
  \caption{Nodes of $\mathcal{S}_1$}
  \label{fig:node1_tdc}
\end{subfigure}
\begin{subfigure}{0.49\columnwidth}
\centering
  \includegraphics[width=.9\linewidth]{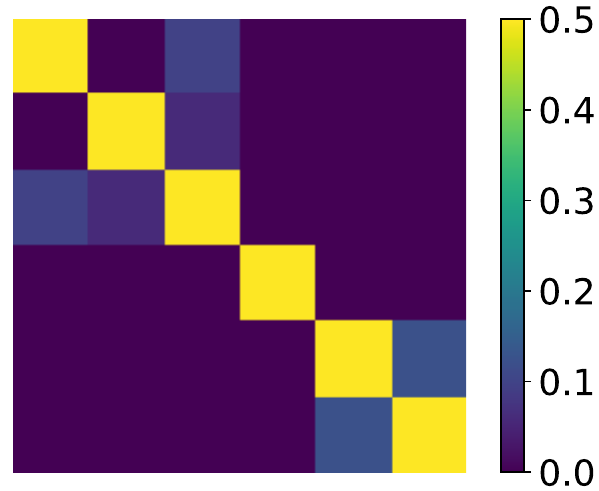}
  \caption{Edges of $\mathcal{S}_2$}
  \label{fig:edge2_tdc}
\end{subfigure}
\begin{subfigure}{0.49\columnwidth}
\centering
  \includegraphics[width=.9\linewidth]{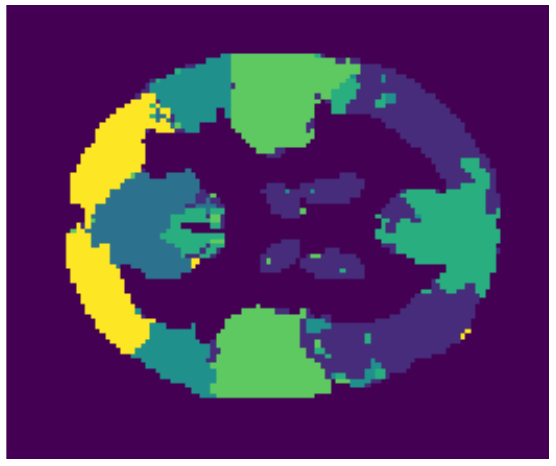}
  \caption{Nodes of $\mathcal{S}_2$}
  \label{fig:node2_tdc}
\end{subfigure}
\begin{subfigure}{1.\columnwidth}
\centering
  \includegraphics[width=0.15\linewidth]{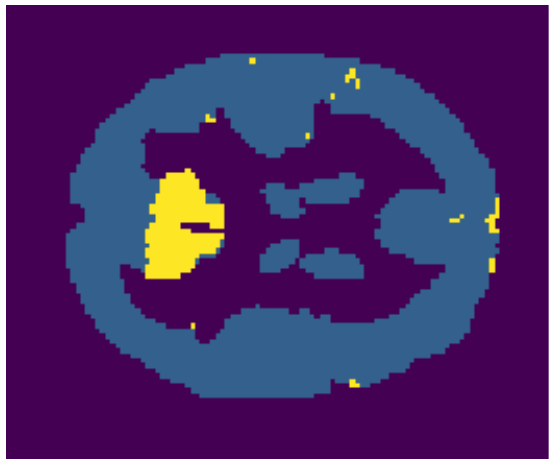}
  \includegraphics[width=0.15\linewidth]{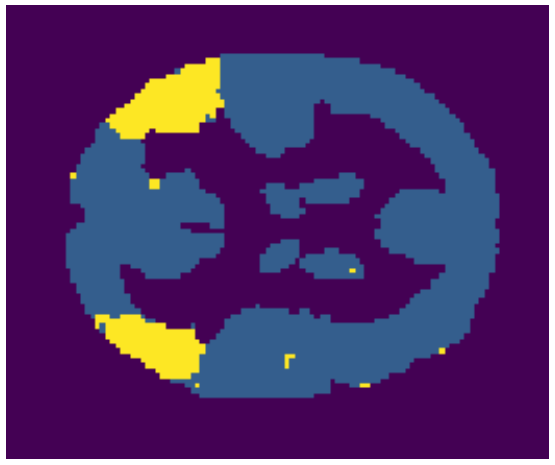}
  \includegraphics[width=0.15\linewidth]{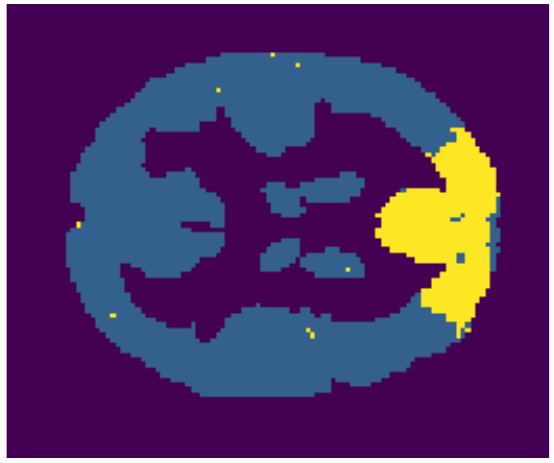}
  \includegraphics[width=0.15\linewidth]{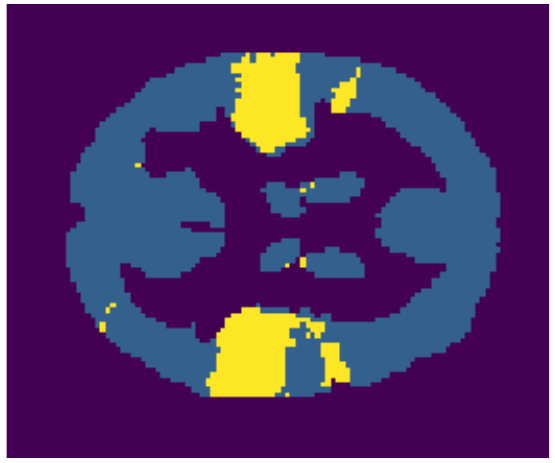}
  \includegraphics[width=0.15\linewidth]{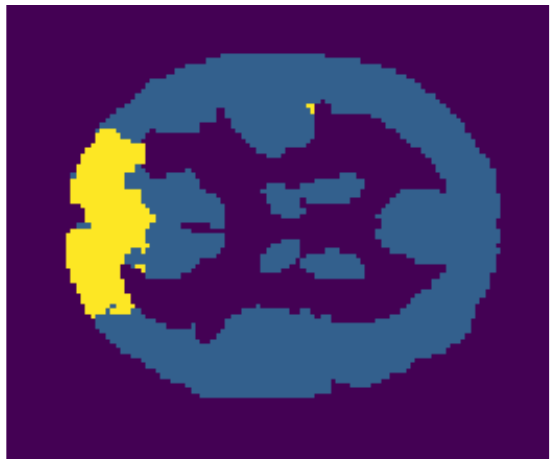}
  \includegraphics[width=0.15\linewidth]{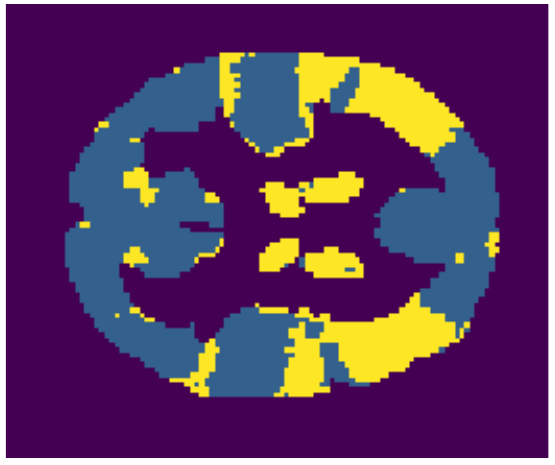}
  \caption{Nodes (1-6) of $\mathcal{S}_1$}
  \label{fig:ind1_tdc}
\end{subfigure}
\begin{subfigure}{1.\columnwidth}
\centering
  \includegraphics[width=0.15\linewidth]{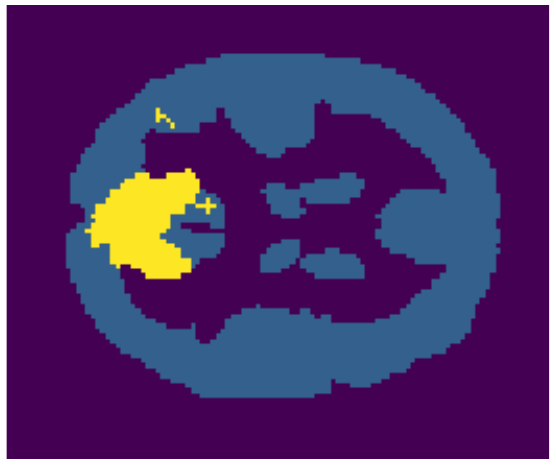}
  \includegraphics[width=0.15\linewidth]{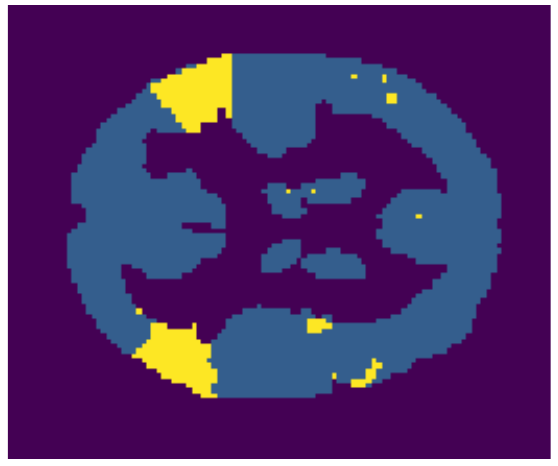}
  \includegraphics[width=0.15\linewidth]{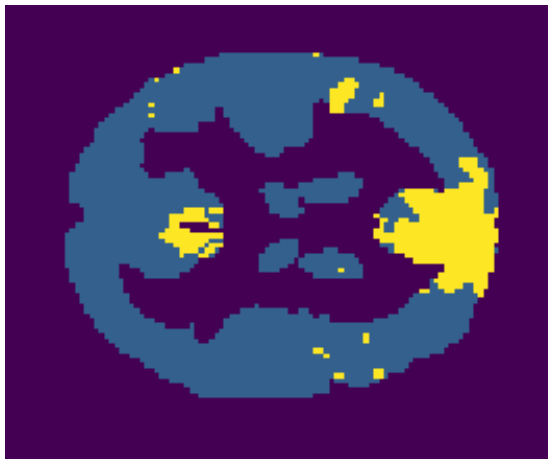}
  \includegraphics[width=0.15\linewidth]{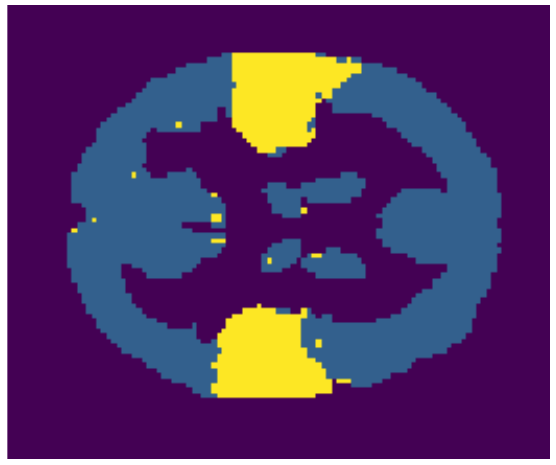}
  \includegraphics[width=0.15\linewidth]{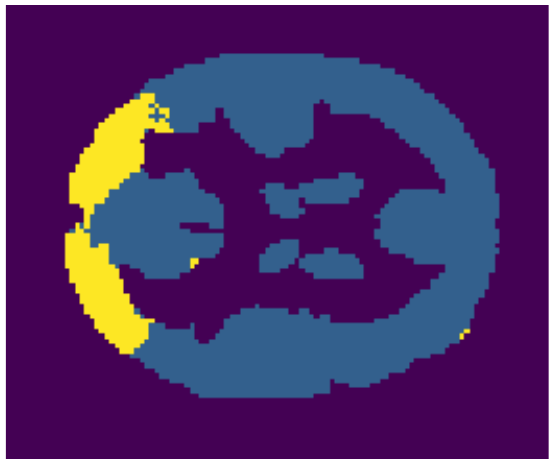}
  \includegraphics[width=0.15\linewidth]{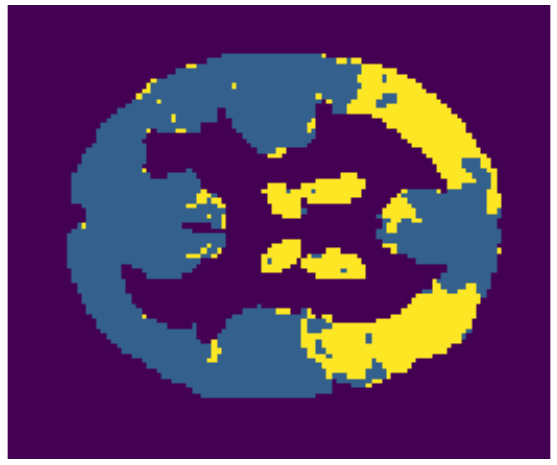}
  \caption{Nodes (1-6) of $\mathcal{S}_2$}
  \label{fig:ind2_tdc}
\end{subfigure}
\vspace{-15pt}

\caption{Discovered results of multi-state brain network in TDC subjects ($k=6$).}
\label{fig:tdc1}
\vspace{-15pt}
\end{figure}